\documentclass[twocolumn,letterpaper]{IEEEAerospaceCLS}  

\usepackage[numbers]{natbib}
\usepackage{multicol}
\usepackage[bookmarks=true]{hyperref}

\usepackage{graphicx}
\usepackage[ruled, vlined, linesnumbered]{algorithm2e}
\usepackage{url}            
\usepackage{amsfonts}       
\usepackage{nicefrac}       
\usepackage{microtype}      
\usepackage{xcolor}
\usepackage{bbm}
\usepackage{bm}
\usepackage{tabularx}
\usepackage{tikz,color,chngpage,soul,csquotes,floatrow,yfonts}
\usepackage{multirow}
\usepackage{subcaption}
\usepackage{tcolorbox}
\usepackage{etoolbox}
\usepackage{mdframed} %
\usepackage{listings}%
\usepackage{enumitem}
\usepackage{booktabs}       
\usepackage{dsfont, amsmath}
\usepackage{cleveref}

\AtBeginEnvironment{tcolorbox}{\scriptsize}

\begin{document}
\title{Space-LLaVA: a Vision-Language Model Adapted to Extraterrestrial Applications}

\author{
Matthew Foutter\textsuperscript{1}, Daniele Gammelli\textsuperscript{2}, Justin Kruger\textsuperscript{2}, Ethan Foss\textsuperscript{2}, Praneet Bhoj\textsuperscript{3}, Tommaso Guffanti\textsuperscript{2},\\
Simone D'Amico\textsuperscript{2}, Marco Pavone\textsuperscript{2,4}\\
\\
Stanford University\\
450 Jane Stanford Way\\
Stanford, CA 94305\\
\\
\{mfoutter; gammelli; jjkruger; erfoss; praneet; tommaso; damicos; pavone\}@stanford.edu\\
\thanks{\textsuperscript{1} Dept. of Mechanical Engineering: 440 Escondido Mall, Stanford, CA 94305; \textsuperscript{2} Dept. of Aeronautics and Astronautics: 496 Lomita Mall, Stanford, CA 94305; \textsuperscript{3} Dept. of Computer Science: 353 Jane Stanford Way, Stanford, CA 94305; \textsuperscript{4} NVIDIA Corporation: 2788 San Tomas Express Way, Santa Clara, CA 95051.}
\thanks{\footnotesize 979-8-3503-5597-0/25/$\$31.00$ \copyright2025 IEEE}
}

\maketitle

\thispagestyle{plain}
\pagestyle{plain}

\begin{abstract}
Foundation Models (FMs), e.g., large language models, possess attributes of intelligence \cite{wei2022cot} which offer promise to endow a robot with the contextual understanding necessary to navigate complex, unstructured tasks in the wild.
We see three core challenges in the future of space robotics that motivate building an FM for the space robotics community: 1) \textit{Scalability} of ground-in-the-loop operations; 2) \textit{Generalizing} prior knowledge to novel environments; and 3) \textit{Multi-modality} in tasks and sensor data. As a first-step towards a \emph{space foundation model}, we programmatically augment three extraterrestrial databases with fine-grained language annotations inspired by the sensory reasoning necessary to e.g., identify a site of scientific interest on Mars, building a synthetic dataset of visual-question-answer and visual instruction-following tuples. We fine-tune a pre-trained LLaVA 13B \cite{Liu2023ImprovedBW} checkpoint on our augmented dataset to adapt a Vision-Language Model (VLM) to the visual semantic features in an extraterrestrial environment, demonstrating FMs as a tool for \emph{specialization} and enhancing a VLM's zero-shot performance on \emph{unseen} task types in comparison to state-of-the-art VLMs. Ablation studies show that fine-tuning the language backbone and vision-language adapter in concert is key to facilitate adaption while a small percentage, e.g., 20\%, of the pre-training data can be used to safeguard against catastrophic forgetting.

\end{abstract}

\tableofcontents

\section{Introduction}

Advancements in the development of internet-scale Machine Learning (ML) models trained through self-supervision on a corpus of human knowledge, i.e., Foundation Models (FMs) \cite{bommasani2021opportunities}, provide an opportunity to automate complex decision making and reasoning transcribed through language, video, and speech. 
State-of-the-art (SoTA) Large Language Models (LLMs) empirically demonstrate strong commonsense reasoning \cite{wei2022cot, wei2022finetuned, Kojima2022LargeLM} and semantic understanding \cite{sinha2024realtime, Elhafsi2023SemanticAD, pgvlm2024} that, for example, enable them to serve as runtime monitors \cite{Elhafsi2023SemanticAD, sinha2024realtime} and language-based planners \cite{Lin2023, ren2023robots} for long-horizon tasks in robotics. These commonsense reasoning capabilities make the use of FMs attractive in space robotics, satellite operations, and other space-related domains, where these models show the potential to mitigate core challenges in the field such as: 1) \textit{Scalability} of Ground-in-the-Loop (GITL) operations; 2) \textit{Generalizing} prior knowledge to novel environments; and 3) \textit{Multi-modality} in tasks and sensor data.
\begin{figure}[t!]
    \centering
    \includegraphics[width=\linewidth]{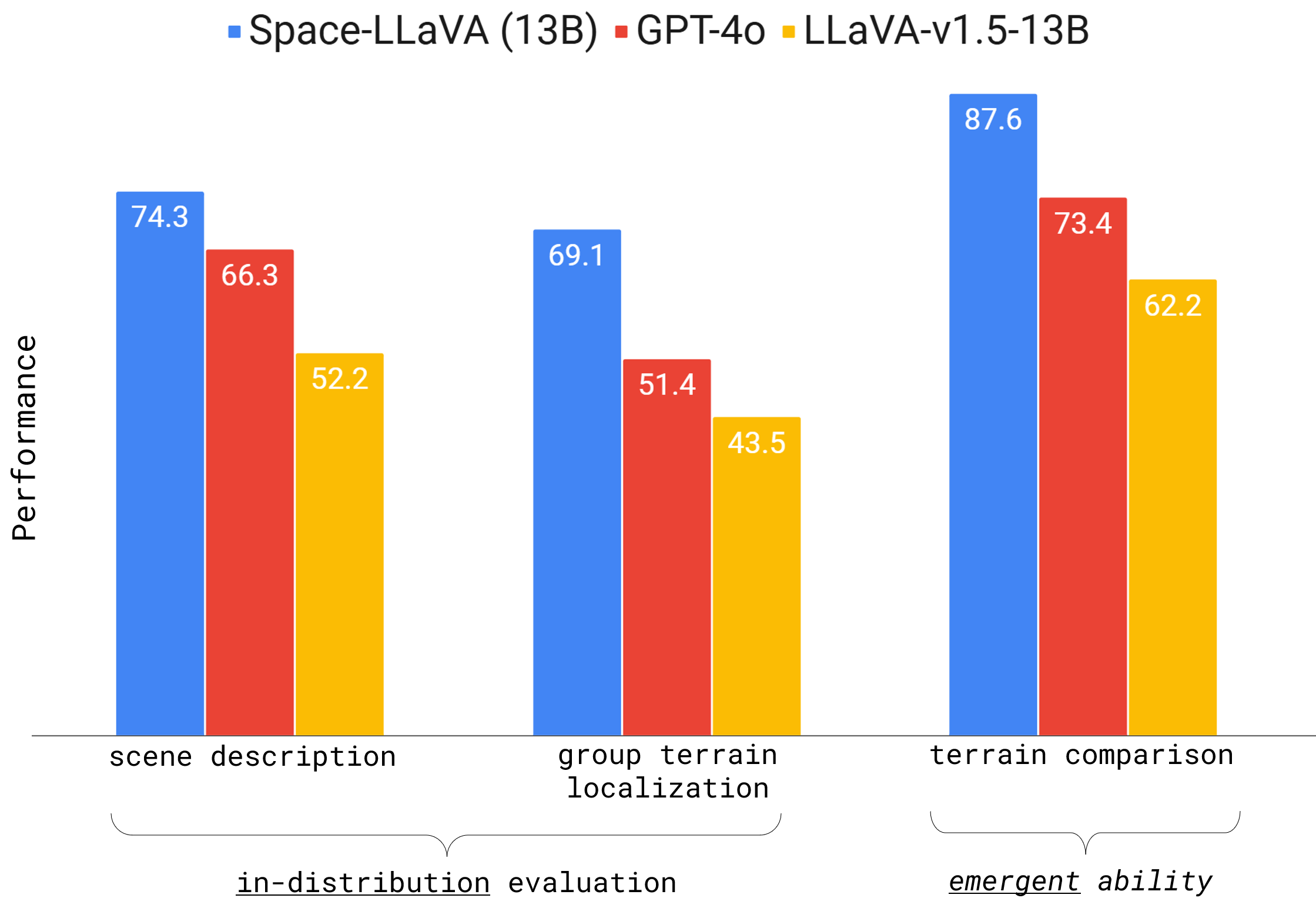}
    \caption{{\fontfamily{cmtt}\selectfont Space-LLaVA} outperforms SoTA VLMs, e.g., GPT-4o \cite{openai_gpt4o} and base LLaVA \cite{Liu2023ImprovedBW}, annotating withheld observations from our synthetic dataset of extraterrestrial, planetary imagery and learns to service queries on an unseen task type as an \emph{emergent ability}.}
    \label{fig:space-llava-performance-evaluation}
\end{figure}

\begin{figure*}[ht!]
    \centering
    \includegraphics[width = \linewidth]{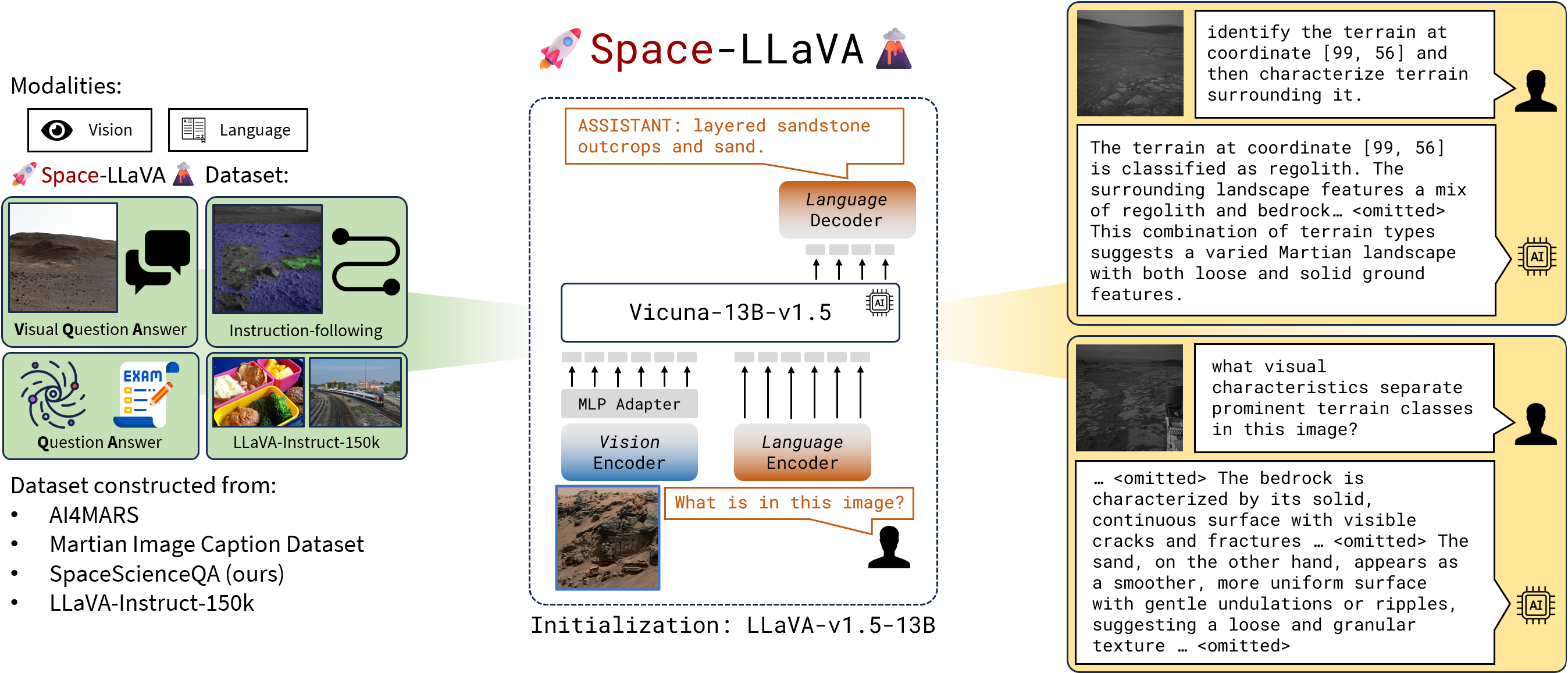}
    \caption{We present {\fontfamily{cmtt}\selectfont Space-LLaVA}, initialized from a pre-trained LLaVA 13B model \cite{Liu2023ImprovedBW} and fine-tuned to extraterrestrial applications with our synthetically generated dataset of, e.g., instruction-following, conversations constructed from three extraterrestrial datasets. This model accepts two data modalities: RGB images and text. Each image is mapped into a shared latent space by the model's image encoder and multi-modal adapter from which a large language model produces a response in natural language. As such, our general-purpose model can be used, among other tasks, as a tool for language annotation servicing requests previously withheld from training.}
    \label{fig:spacellava}
\end{figure*}

\textbf{Ground-in-the-Loop (GITL) Operations}: Recent extraterrestrial robotic missions have operated a single spacecraft in a tightly scheduled GITL paradigm, which scales poorly and is expensive, e.g., the operational cost for the Perseverance rover was almost \$300 million \cite{Perseverance}. As efforts advance toward more cost-effective multi-robot missions \cite{Artemis}, FMs offer opportunities to assist human supervisors by, e.g., summarizing observations, diagnosing issues, setting higher-level robot objectives, or coordinating a human-robot interface \cite{royce2024enablingnovelmissionoperations}. More generally, there has been immense growth in usage of the space environment in recent years, with the deployment of satellite mega-constellations and the pursuit of ambitious new objectives such as in-orbit servicing, assembly, and manufacturing (ISAM). This growth necessitates new paradigms for space mission operations that uphold the quality and safety standards of traditional GITL methods while enabling autonomous and responsive decision-making both onboard spacecraft and within ground systems.

\textbf{Generalizing Prior Knowledge}: Space exploration is inherently zero-shot, characterized by novel, unstructured, and under-mapped environments. Therefore, there is a growing need for models capable of generalizing knowledge from prior missions. For instance, computer vision navigation techniques developed for Mars missions could be adapted for the Artemis moon mission \cite{MatthiesEtAl2022}, insights from Ingenuity’s flights on Mars may inform the Dragonfly mission on Titan \cite{dragonfly}, and data from previous satellite rendezvous missions could support rendezvous and servicing of new in-space targets. 
Generalist FMs have the potential to harness both their extensive pre-training corpus and prior mission data to enable robust and reliable inference in these uncertain and novel environments. 

\textbf{Multiple Sensor Modalities}: Diverse robot embodiments in space offer a wealth of sensing modalities beyond standard encodings such as language and vision. A satellite may combine a gyroscope, star tracker, and Global Navigation Satellite Systems (GNSS) data for attitude determination and orbit determination in Earth orbit, while a Mars rover may use stereo camera and radar technology to build a map of the environment. Therefore, the development of FMs that jointly encode \textit{all} of a robot's or satellite's sensing modalities into a shared representation space may result in generalist capabilities that can be applied to any chosen mission architecture. 

Accordingly, this paper presents a preliminary investigation into pre-trained multi-modal FMs as applied to extraterrestrial robotics, focusing on techniques aimed at achieving the three aforementioned goals. 
This, however, is immediately met with four key challenges. First, existing FMs are trained primarily, if not entirely, on terrestrial image captioning and Visual-Question-Answer (VQA) datasets \cite{tsung2014mscoco, antol2015vqa}. Hence, these models are prone to hallucinations, e.g., producing inscrutable or inaccurate responses, when applied to observations from extraterrestrial applications. In this work, we investigate whether SoTA FMs, such as GPT-4o \cite{openai_gpt4o}, can be applied to extraterrestrial observations in a zero-shot manner or require adaptation through fine-tuning to address potential visual domain and knowledge gaps. Second, in robotics---particularly in space robotics---it is essential to develop learning algorithms that can perform effectively despite limited data availability. Moreover, extraterrestrial datasets annotated with semantic features in natural language are exceedingly rare, if they exist at all. Therefore, adapting the generalist prior of a language-based FM to space robotics demands cost-effective, high-quality, and scalable techniques to generate large databases of language-based Question-Answer (QA) pairs from existing small-scale datasets like PDS \cite{NASA_PDS}, AI4Mars \cite{ono2021ai4mars}, or SPEED+ \cite{park2022speedp}, which are in themselves insufficient for large FM development. Third, we must design meticulous and targeted training tasks that elicit the fine-grained, sensory reasoning capabilities required to apply FMs towards the three aforementioned goals. These tasks should go beyond generic descriptions of an observation to address the specific demands of space robotics. Fourth, in addressing these challenges, we must remain cognizant of the degree to which our contributions, e.g., fine-tuning, stain the generalist prior of an FM: care must be taken to prevent catastrophic forgetting, ensuring that the model retains its broad utility while acquiring the specialized knowledge needed for extraterrestrial applications.

As a first step towards a \emph{space foundation model}, we demonstrate the opportunity for FMs to mitigate data scarcity by synthetically augmenting extraterrestrial science datasets, such as AI4Mars. Specifically, we generate a multi-modal dataset comprised of 150k QA tuples designed to emulate the detailed sensory reasoning required for tasks like identifying sites of scientific interest. We fine-tune an open-source Vision-Language Model (VLM) on our synthetic dataset, herein referred to as the {\fontfamily{cmtt}\selectfont Space-LLaVA dataset}, and demonstrate the model's utility by providing language annotations on planetary observations and tasks withheld from training. That is, {\fontfamily{cmtt}\selectfont Space-LLaVA} which we train, among other objectives, to perform data curation, e.g., {\fontfamily{cmtt}\selectfont scene description}, can provide high-quality language annotations on Martian imagery for specialized ML algorithms, and, akin to research in instruction-tuning \cite{wei2022finetuned}, we find that the model can proficiently answer previously unseen annotation requests as an \emph{emergent ability} in comparison to current SoTA VLMs. We visually present {\fontfamily{cmtt}\selectfont Space-LLaVA} in \Cref{fig:spacellava} along with a brief characterization of our {\fontfamily{cmtt}\selectfont Space-LLaVA dataset} and two example generations from the model.

\begin{figure}[t]
    \begin{subfigure}[t]{0.47\linewidth}
        \centering
        \includegraphics[width=\linewidth]{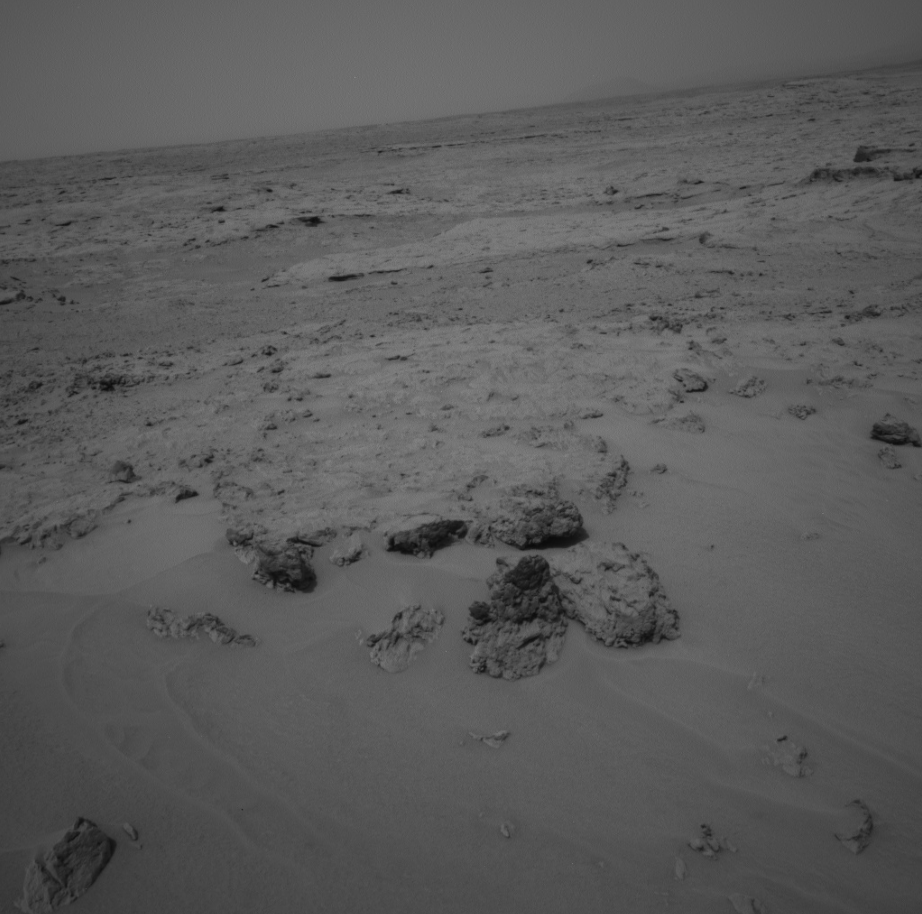}
        \caption{MSL NAVCAM image of Mars' landscape.}
        \label{fig:raw-mars-terrain}
    \end{subfigure}
    \hfill
    \smallskip
    \begin{subfigure}[t]{0.47\linewidth}
        \centering
        \includegraphics[width=\linewidth]{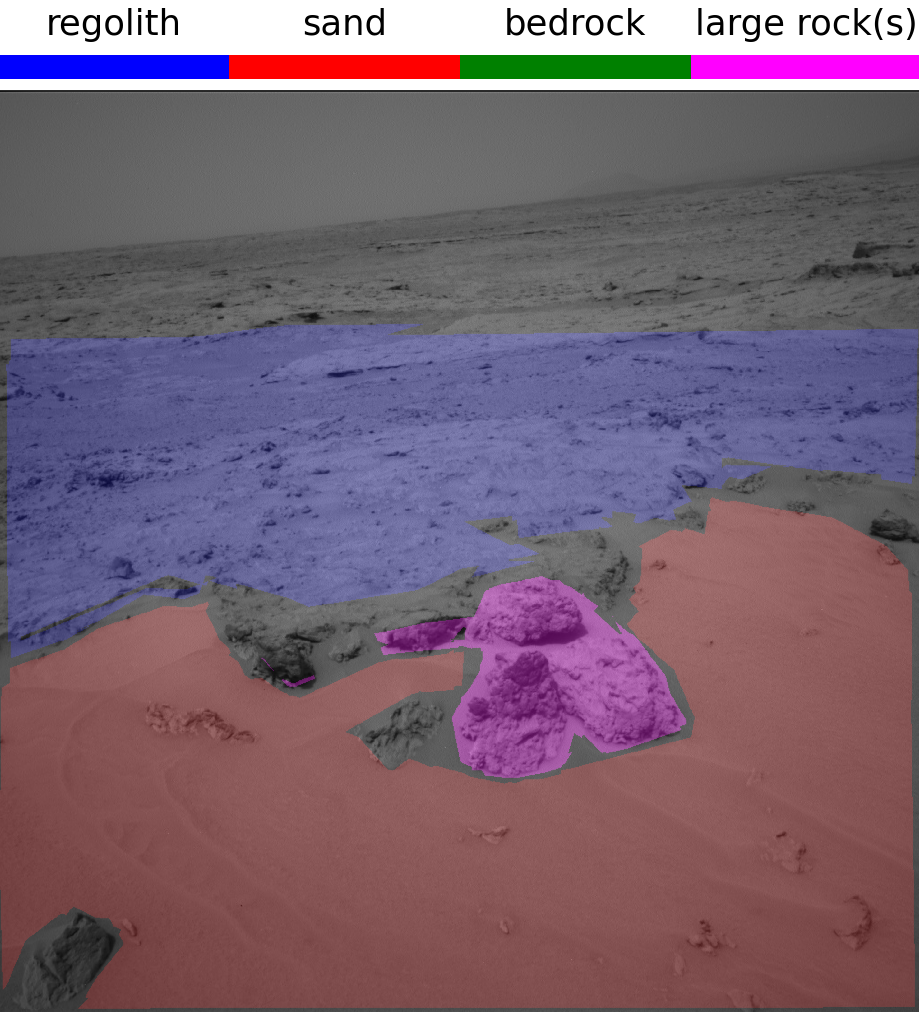}
        \caption{Crowd-sourced segmentation masks superimposed on Martian landscape.}
        \label{fig:segment-mars-terrain}
    \end{subfigure}
    \caption{The AI4Mars dataset \cite{ono2021ai4mars} provides access to image captures of Mars' terrain with crowd-sourced annotations for four terrain classes: ``regolith", ``sand", ``bedrock", ``large rock(s)". Terrain beyond 30m is left unlabeled.}
    \label{fig:ai4mars}
\end{figure}

Additionally, we safeguard against catastrophic forgetting---the phenomenon by which an ML algorithm's broad performance degrades outside of the scope captured by the fine-tuning dataset \cite{zhai2023investigating}---by co-training {\fontfamily{cmtt}\selectfont Space-LLaVA} with a percentage of the model's pre-training instruction-following dataset, i.e., LLaVA-Instruct-150k \cite{liu2023visual}. We validate that our training regime conserves an FM's generalist prior through a holistic evaluation on standard VQA benchmarks; for completeness, we characterize {\fontfamily{cmtt}\selectfont Space-LLaVA's} trustworthiness out-of-domain on the visually sparse and distinct domain of on-orbit imagery, documenting the competency of current FMs on orbital data and the extent to which these models can augment satellite operators.

Lastly, we explore the potential of integrating FMs into modular autonomy stacks, where an FM interfaces with diverse lower- and higher-level components.
Specifically, we leverage a simulator platform that provides realistic 3D environments for testing and validating autonomous rover operations on the lunar surface and demonstrate the use of an FM as a high-level path planner and runtime monitor for a lunar rover.

Our evaluations demonstrate that: 1) existing VLMs are deficient visual reasoners in extraterrestrial applications; 2) our {\fontfamily{cmtt}\selectfont Space-LLaVA dataset} endows a SoTA VLM with zero-shot performance increases servicing unseen extraterrestrial task types through instruction-tuning; 3) a small percentage, e.g., 20\%, of the pre-training data is sufficient to safeguard against catastrophic forgetting; 4) FMs can be effectively integrated into modular autonomy stacks to enable embodied high-level planning in space robotics.



\section{Related Work}

\textbf{Vision-Language Models}: The advent of the Transformer \cite{vaswani2017transformer} and derivative architectures, e.g., the Vision-Transformer \cite{dosovitskiy2021vit}, have powered recent advances in natural language and image processing through the use of VLMs trained on internet-scale text and image databases, e.g., Common Crawl and WebImageText \cite{srinivasan2021wit}. Early work in vision-language modeling at scale \cite{radford2021clip} aligns a latent representation of vision and language by using a vision and text encoder with a contrastive learning objective; a VLM builds on this architecture by using a language model for open-ended visual reasoning such as VQA \cite{Li2023BLIP2,dai2023instructblip, liu2023visual, Driess2023PaLMEAE}. In this work, we investigate adapting LLaVA-v1.5-13B \cite{Liu2023ImprovedBW} to extraterrestrial robotics through fine-tuning given this model is SoTA among open-source models on standard VQA benchmarks \cite{antol2015vqa, Hudson2019GQAAN}.

\begin{figure}[b]
    \centering
    \includegraphics[width=0.6\linewidth]{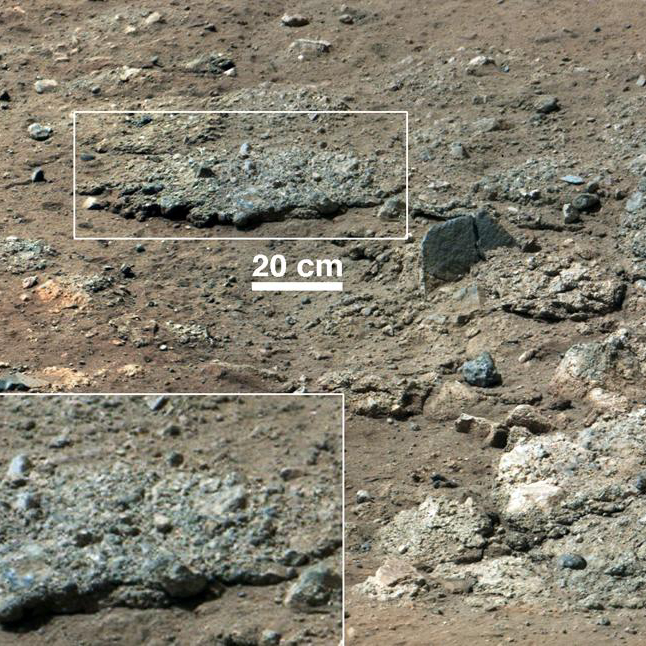}
    \caption{Sample from the Martian Image Caption Dataset (MICD): ``conglomerate outcrops and float rocks and regolith."}
    \label{fig:micd-sample}
\end{figure}

\textbf{Foundation Models in Robotics}: Prior work has incorporated foundation models within the broader robot autonomy stack in various ways ranging from planning \cite{Lin2023}, decision making \cite{pmlr-v229-zitkovich23a} and semantic reasoning \cite{Elhafsi2023SemanticAD, sinha2024realtime} to visual reasoning \cite{shah2023vint}. However, the opportunity for foundation models in extraterrestrial robotics represents an emerging area of research. The Robot Operating System Agent \cite{royce2024enablingnovelmissionoperations} employs FMs to build a human-robot language interface for operators using bespoke robotic technology; SpaceTransformers \cite{spacetransformers} fine-tunes three variations of the BERT \cite{bert} architecture on a corpora of systems engineering texts and an augmented mission standards dataset to recognize space mission requirements. In a similar context, SpaceQA \cite{spaceqa} builds on SpaceTransformers by creating an LLM for space mission design, which is suitable for pre-launch mission design and evaluation but is not extensible to in-flight robot operations. Toward the use of a foundation model for in-flight operation, \cite{rodriguezfernandez2024language} leverages GPT-3.5 \cite{brown2020gpt3} as the policy backbone for language-based autonomous satellite operations in the Kerbal Space Program Differential Games Challenge. We aim to extend this work by incorporating both vision and language into a shared representation for enhanced reasoning and compatibility with a broader suite of extraterrestrial robot embodiments.

\textbf{Large-scale Dataset Curation}: Existing work in large-scale robot data collection is exemplified by Open X-Embodiment \cite{Padalkar2023OpenXR} representing the largest open-source documentation of robot manipulation to date. In \cite{pgvlm2024}, the authors develop the PhysObject dataset annotated with the physical properties of common objects. While this work largely uses automatic annotation in data collection, the PhysObject dataset is itself an image classification dataset, e.g., predict the mass of a cup. We aim to extend this work by developing an extraterrestrial dataset for \textit{visual reasoning}. \cite{Marcu2023LingoQAVQ, ma2023dolphins} curate a large-scale autonomous driving VQA benchmark to enable perception, prediction, and planning; however, they require language annotations from human operators, which is likely incompatible with long-horizon data collection at scale. Consequently, our work is distinguished from existing work by programmatically generating a dataset of language annotations for visual reasoning on which to fine-tune a VLM in the context of extraterrestrial robotics.

\section{Architecting Space-LLaVA}\label{sec:problem_formulation}

In this work, we fine-tune an open-source VLM on our {\fontfamily{cmtt}\selectfont Space-LLaVA dataset}, which is synthetically generated by augmenting three extraterrestrial databases: AI4Mars \cite{ono2021ai4mars}, Martian Image Caption Dataset (MICD) \cite{qiu2020scoti} and SpaceScienceQA (ours). The Mars Science Laboratory (MSL) subset of the AI4Mars dataset encompasses 17k images with crowd-sourced segmentation masks of Mars' terrain gathered from the Curiosity rover. A representative example of raw terrain from the MSL AI4Mars dataset and its associated semantic masks for each terrain class is provided in \Cref{fig:ai4mars}. Further, the MICD currently offers 3k image caption pairs derived through crowd-sourced annotations in natural language. We provide a representative image sample and caption from the MICD in \Cref{fig:micd-sample}. Finally, we defer the discussion of our SpaceScienceQA dataset to the proposed approach in \Cref{sec:space-science-QA} and offer two representative samples from the SpaceScienceQA dataset in \Cref{fig:space-science-qa-sample}.

We require a high-quality and scalable technique to augment these extraterrestrial datasets with complex, semantic reasoning transcribed in natural language as, e.g., VQA, conversations to serve as the foundation for fine-tuning. Through fine-tuning, we aim to endow a VLM with the crowd-sourced knowledge base originally collected by these datasets and demonstrate the model's utility as a tool for continued data annotation as an \emph{emergent ability}. That is, {\fontfamily{cmtt}\selectfont Space-LLaVA} should be useful as a general-purpose tool for providing language annotations on extraterrestrial data in response to queries beyond the fine-tuning dataset's limited scope.

We ground a VLM in the visual and semantic features of our {\fontfamily{cmtt}\selectfont Space-LLaVA dataset} by fine-tuning LLaVA-v1.5-13B \cite{Liu2023ImprovedBW} on our augmented dataset with the standard auto-regressive language modeling loss. Suppose we curate a dataset $D = \{(\textbf{I}^{(i)}, \textbf{Q}^{(i)}, \textbf{A}^{(i)})\}_{i=1}^n$ consisting of $n$ image $\textbf{I}^{(i)} \in \mathbb{R}^{\textit{h} \times \textit{w} \times \text{3}}$, question $\textbf{Q}^{(i)} \in \mathbb{R}^{T_Q}$, and answer $\textbf{A}^{(i)} \in \mathbb{R}^{T_A}$, tuples where $T_Q$ and $T_A$ denote the maximum tokenized question and answer sequence length, respectively, with padding. 

We fine-tune LLaVA-v1.5-13B by freezing certain parameters in the model, e.g., only fine-tuning the language backbone, to optimize the objective
\begin{equation} \min_{\hat{\theta} \subseteq \Theta} \ L(\hat{\theta} \mid \mathcal{D}) \label{opt_statement}, \end{equation}
where we construct $L(\hat{\theta} \mid \mathcal{D})$ as the negative log-likelihood loss on token generation assuming samples are independent and identically distributed and using the chain rule factorization for auto-regressive generation.
More formally, we define:
\begin{equation} L(\hat{\theta} \mid {D}) = -\frac{1}{n}\sum_{i=1}^n\sum_{t=1}^T \log p_{\hat{\theta}}(x_{t+1}^{(i)} \mid \textbf{I}^{(i)}, \textbf{Q}^{(i)}, \textbf{A}^{(i)}_{1:t}) \label{language_loss}, \end{equation}
where each term in the summation represents the log-likelihood, under the model's current weights $\hat{\theta}$, to predict the ground-truth next text token in the answer sequence $\textbf{A}^{(i)} = \{x_1^{(i)}, x_2^{(i)}, \hdots, x_T^{(i)}\}$ conditioned on the sample's visual input, associated question and the full answer sequence preceding $x_{t+1}^{(i)}$. 
Here, $\hat{\theta} \subseteq \Theta$ indicates that the unfrozen weights are a subset of the model's weights $\Theta$. We evaluate the quality of the fine-tuned model's responses in comparison to a base model by prompting GPT-4 \cite{OpenAI_GPT4_2023} to score each response on a numeric scale and choose which response is preferable relative to the query's ground-truth answer. A template of the prompt we provide to GPT-4 is provided in \Cref{sec:appendix_pref_analysis} with further discussion on the prompt's construction and evaluation pipeline.

That is, through fine-tuning a VLM on our synthetic dataset with semantic annotations, we measure whether the fine-tuned model outperforms SoTA VLMs on the same task without adaptation, i.e., zero-shot.

\section{Data Generation Pipeline}\label{sec:proposed_approach}

\begin{figure}[t]
    \centering
    \includegraphics[width=0.75\linewidth]{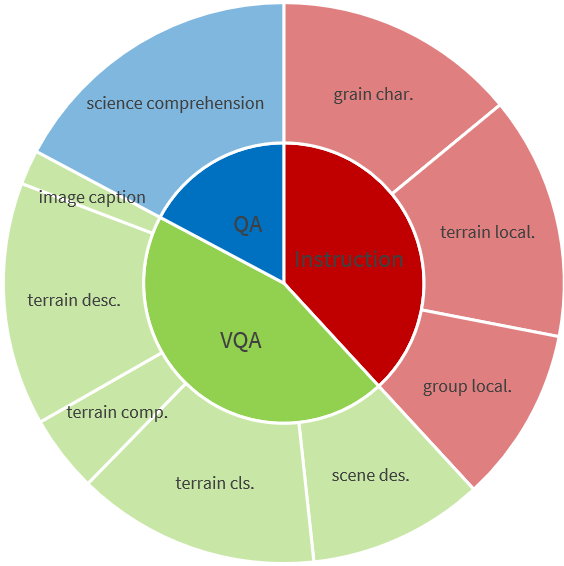}
    \caption{The proportional representation of prompt style, e.g., instruction-following, and the designed fine-tuning tasks, e.g., {\fontfamily{cmtt}\selectfont grain characterization}, in our {\fontfamily{cmtt}\selectfont Space-LLaVA dataset}. All instruction and VQA-based tasks are derived from the AI4Mars \& MICD datasets, while the {\fontfamily{cmtt}\selectfont SpaceScienceQA dataset} represents the only language QA-based category.}
    \label{fig:spacellava-composition}
\end{figure}

The scarcity of high-quality, extraterrestrial data represents a persistent challenge in the pursuit of artificial intelligence within the space robotics community. Further, while aforementioned databases, e.g., AI4Mars \cite{ono2021ai4mars}, SPEED+ \cite{park2022speedp}, offer extraterrestrial data with high-quality labels, e.g., semantic segmentation masks and satellite pose, these multi-modal databases lack annotations in natural language which may be leveraged for adaptation by a traditional FM. Therefore, in order to transfer the zero-shot learning capabilities of FMs elicited through instruction-tuning \cite{liu2023visual, wei2022finetuned} to extraterrestrial applications, we develop a VQA generation pipeline based on the AI4Mars and MICD \cite{qiu2020scoti} datasets supplemented by recent publications in astrophysics. Explicitly, we translate AI4Mars' segmentation masks into \emph{visual context} for GPT-assisted annotation of seven terrain-based, semantic tasks on Martian imagery, and inspired by cosmosage \cite{dehaan2024cosmosagenaturallanguageassistantcosmologists}, we introduce our own QA dataset reflecting scientific insights and facts captured by publications in arXiv's astrophysics category, e.g., Earth and Planetary Astrophysics, which we refer to as the {\fontfamily{cmtt}\selectfont SpaceScienceQA dataset}. 

We first discuss our simple and scalable methodology to produce fine-grained sensory reasoning tasks on the AI4Mars dataset and MICD. Then, we detail our approach to synthetically generate high-quality science QA pairs for our {\fontfamily{cmtt}\selectfont SpaceScienceQA dataset}. Our full dataset's composition based on prompt style and the designed fine-tuning tasks is presented in \Cref{fig:spacellava-composition}.

\subsection{GPT-assisted Annotation: AI4Mars \& MICD Datasets}\label{sec:ai4mars_scoti}

\begin{figure}[b!]
    \begin{subfigure}{\linewidth}
        \includegraphics[width=\linewidth]{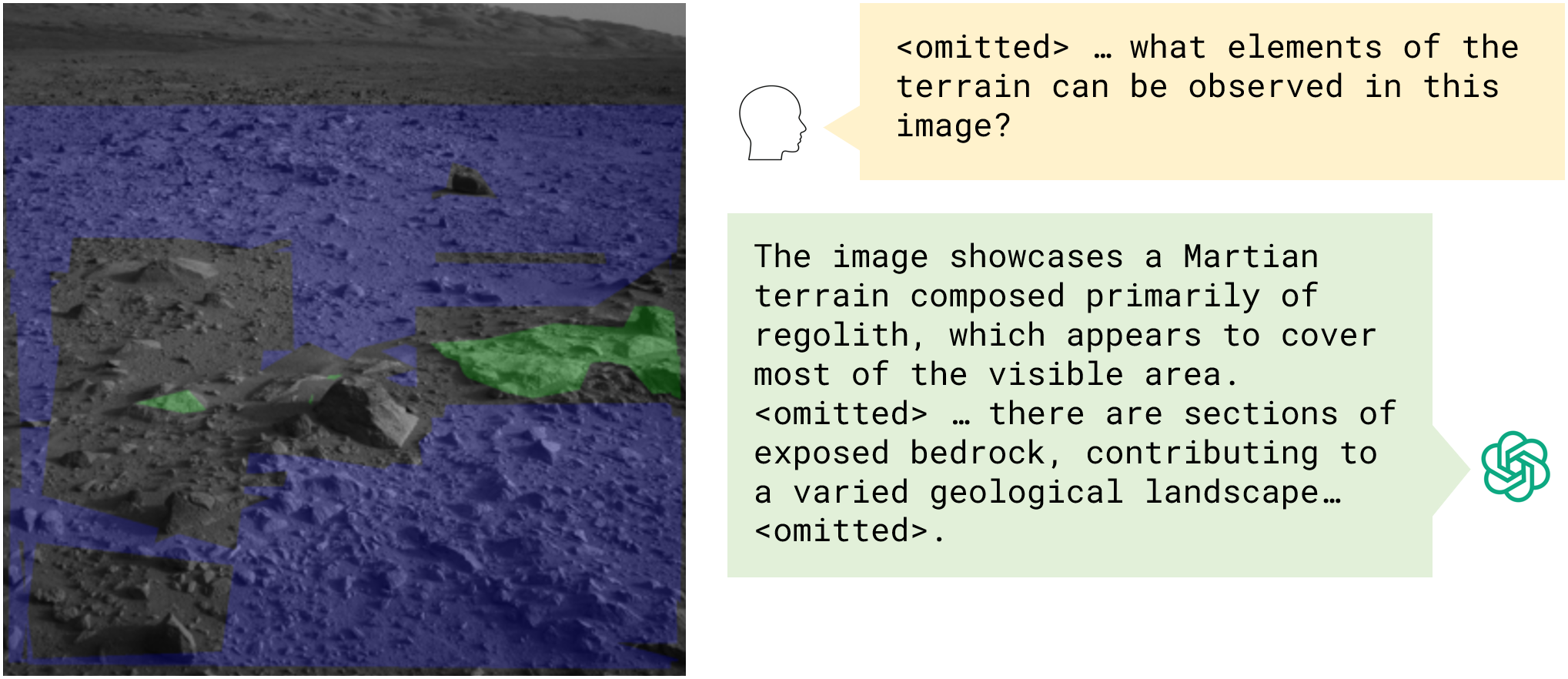}
        \caption{{\fontfamily{cmtt}\selectfont Terrain Description}: GPT-4o annotates a candidate AI4Mars landscape with a description of the terrain in view.}
        \label{fig:ai4mars-terrain-desc}
    \end{subfigure}
    \vfill
    \smallskip
    \begin{subfigure}{\linewidth}
        \includegraphics[width=\linewidth]{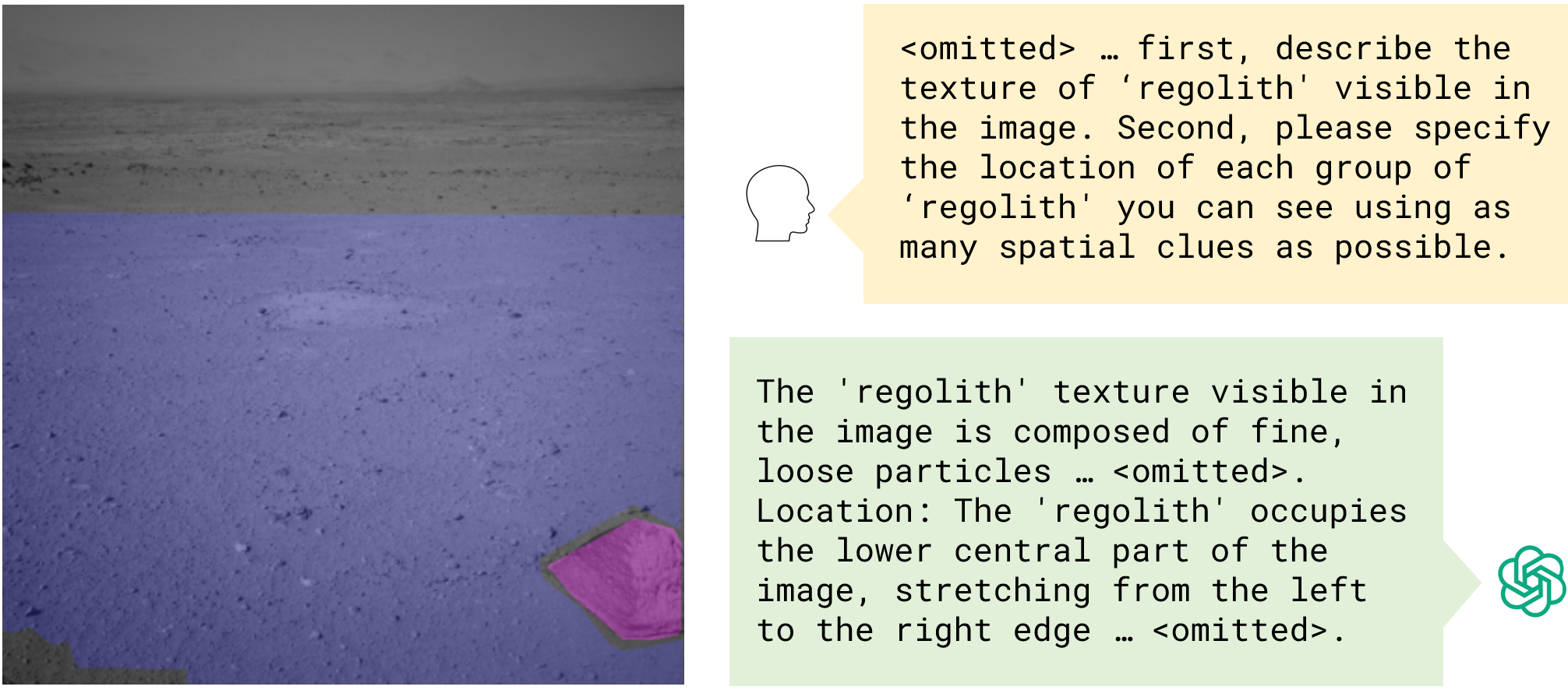}
        \caption{{\fontfamily{cmtt}\selectfont Grain Characterization}: GPT-4o annotates a candidate AI4Mars landscape by detailing the size and arrangement of particles for a particular terrain type.}
        \label{fig:ai4mars-grain-characterization}
    \end{subfigure}
    \caption{Color-coded visual context classifies and localizes terrain in a candidate AI4Mars image through high-quality, semantic segmentation masks. Additional context provided to the model, e.g., associating each terrain type with a unique color, and verbose sections of GPT-4o's response are omitted for brevity.}
    \label{fig:ai4mars_dataset_curation}
\end{figure}

We translate the high-quality, segmentation masks afforded by the AI4Mars dataset, as shown in \Cref{fig:segment-mars-terrain}, into seven distinct, semantic-reasoning tasks through the use of GPT-assisted image annotation. These seven tasks, e.g., {\fontfamily{cmtt}\selectfont terrain comparison}, listed fully in \Cref{sec:appendix_data_augmentation}, are designed to support {\fontfamily{cmtt}\selectfont Space-LLaVA} as a tool for annotating planetary imagery, whose terrain-aware annotations may be used downstream by a specialized, task-specific ML algorithm. For each task, we design a total of ten questions to accomplish the same objective with varied prose, e.g., if the task is {\fontfamily{cmtt}\selectfont scene description}, then we may pose the question as 1) ``describe the landscape in view." or 2) ``what do you see in this image?", etc., so as to discourage over-fitting to a particular prompt's writing style in adaptation, i.e., fine-tuning. Before we query GPT-4o to perform e.g., {\fontfamily{cmtt}\selectfont terrain comparison}, for a particular image, we first superimpose the appropriate terrain segmentation mask(s) on the original MSL NAVCAM image to color-code the landscape, as shown in \Cref{fig:ai4mars_dataset_curation}, creating \emph{visual context} to support GPT-4o's analysis. Through the use of visual context and additional language context provided in the prompt, we request the desired annotation in a format that is readily discernible zero-shot by a SoTA VLM like GPT-4o, i.e., the requested annotation does not require prior, expert knowledge to answer the question. Importantly, all visual and language context is only provided to GPT-4o to promote high-quality data curation; this same context is withheld from training {\fontfamily{cmtt}\selectfont Space-LLaVA} as these features are not available at inference. Further details on the specific prompt used for data curation, e.g., the user and system message, are provided in \Cref{sec:appendix_data_augmentation}.

Then, with the MICD dataset, we have the inverse problem: the MICD dataset provides expert annotations describing geological and terrain features in a candidate Martian image, and we simply must decide the question which appropriately precedes the answer. As before, we design ten questions which request a caption of the image's content, and we provide specific examples in the \Cref{sec:appendix_data_augmentation}.

\begin{table*}[t]
\caption{For each fine-tuned {\fontfamily{cmtt}\selectfont Space-LLaVA} configuration, we report response preference in comparison to SoTA VLMs, e.g., base LLaVA model and GPT-4o. Additionally, we report the average numeric score and standard deviation assigned to the configuration's responses as assessed by GPT-4. The strongest performance for each evaluation metric is highlighted in bold while the best-in-class performance is underlined.}
\centering
\begin{tabular}{cccccccccc}
\toprule
LLaVA-Instruct-150k & & \multicolumn{3}{c}{Frozen} & & \multicolumn{2}{c}{Response Preference} & & Numeric Evaluation\\
\cmidrule{1-1} \cmidrule{3-5} \cmidrule{7-8} \cmidrule{10-10}
Percentage & & VE & MMA & LM & & Ours vs. LLaVA & Ours vs. GPT-4o & & Ours \\ 
\midrule
\parbox[t]{2mm}{\multirow{3}{*}{\rotatebox[origin=c]{90}{90\%}}} & & 
$\checkmark$ & $\checkmark$ & $\times$  & & \underline{91.0\%} & 83.7\% & & \underline{85.9 $\pm$ 9.7\%} \\
& & $\checkmark$ & $\times$ & $\checkmark$  & & 49.8\% & 31.2\% & & 68.6 $\pm$ 15.4\% \\
& & $\checkmark$ & $\times$ & $\times$  & & 90.5\% & \underline{84.3\%} & & 85.8 $\pm$ 10.4\% \\
\midrule
\parbox[t]{2mm}{\multirow{3}{*}{\rotatebox[origin=c]{90}{20\%}}} & & 
$\checkmark$ & $\checkmark$ & $\times$  & & 92.4\% & 85.8\% & & 86.4 $\pm$ 9.4\% \\
& & $\checkmark$ & $\times$ & $\checkmark$  & & 49.1\% & 29.7\% & & 68.5 $\pm$ 15.9\% \\
& & $\checkmark$ & $\times$ & $\times$  & & \textbf{92.9\%} & \textbf{87.1\%} & & \textbf{87.6 $\pm$ 8.5\%}\\
\bottomrule
\end{tabular}
\label{results}
\end{table*}

\subsection{GPT-assisted Annotation: SpaceScienceQA Dataset}\label{sec:space-science-QA}

\begin{figure}[t]
    \centering
    \includegraphics[width=\linewidth]{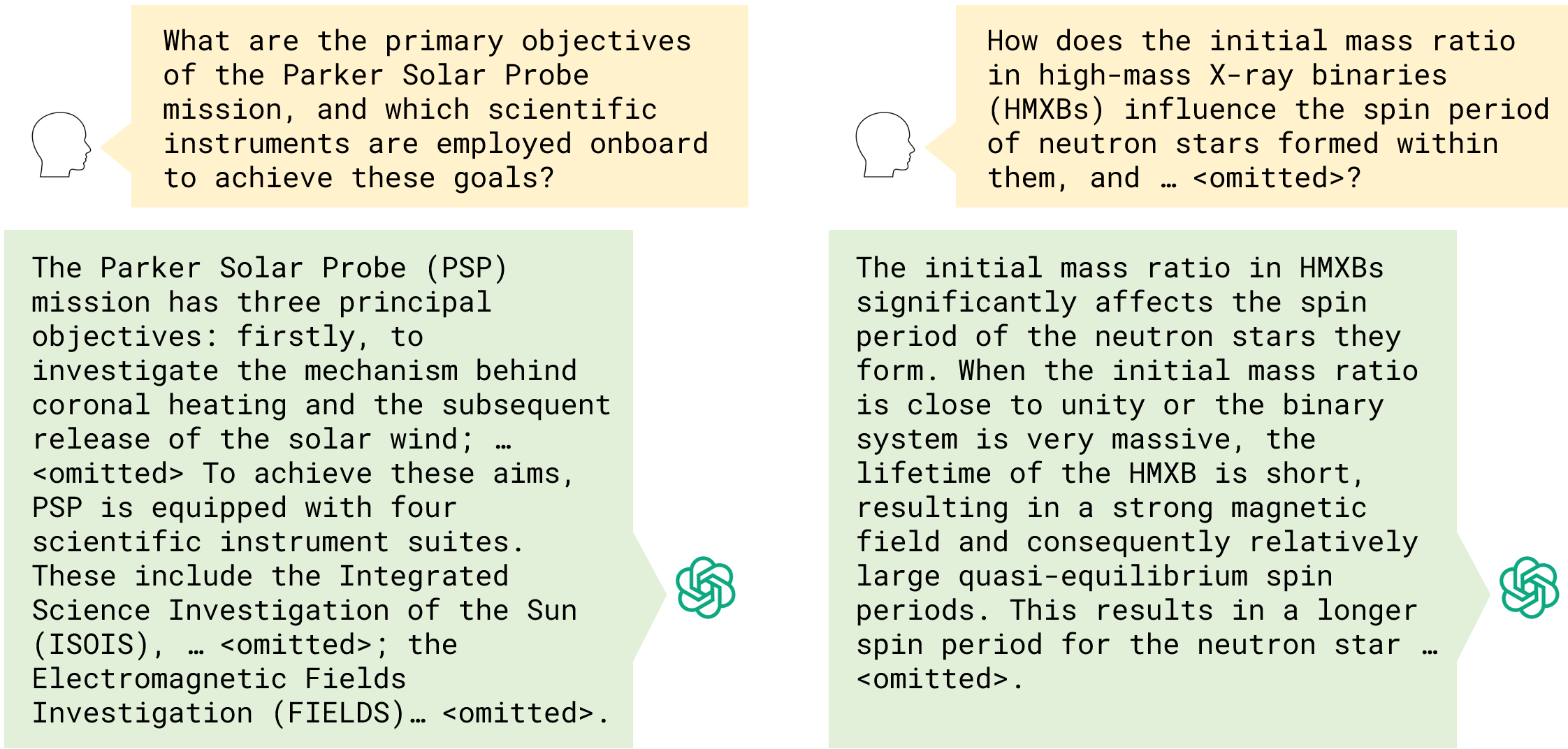}
    \caption{Our {\fontfamily{cmtt}\selectfont SpaceScienceQA dataset} offers QA tuples evaluating a language model's understanding of scientific insights and facts in astrophysics. Verbose sections of the question and answer are omitted for brevity.}
    \label{fig:space-science-qa-sample}
\end{figure}

As a first step to endow an LLM with the understanding of extraterrestrial science and encourage a community effort to build an FM for space robotics, we've designed the {\fontfamily{cmtt}\selectfont SpaceScienceQA dataset}, shown in \Cref{fig:space-science-qa-sample}, capturing scientific insights and facts from 1,000 of the most recent publications in astrophysics. Inspired by cosmosage \cite{dehaan2024cosmosagenaturallanguageassistantcosmologists}, 
we use a SoTA LLM, i.e., GPT-4, to automatically generate a large-scale dataset of 25k science QA pairs designed to evaluate one's understanding of scientific concepts in astrophysics. Central to our approach is a simple and programmatic methodology by which we quantify the \emph{quality} and the potential for \emph{information gain} through fine-tuning for each candidate QA pair. Specifically, similar to existing work, we use GPT-4 to independently judge the fidelity of a QA pair with respect to the original publication, and unlike existing work, we quantify the opportunity for information gain through fine-tuning by evaluating LLaVA's language model, i.e., Vicuna-13B-v1.5 \cite{vicuna2023}, on a candidate QA pair---rejecting QA pairs on which Vicuna-13B-v1.5 is already proficient. Further detail and discussion on the full data generation pipeline and evaluation are contained in \Cref{sec:appendix_scienceQA}. 

\section{Experiments \& Discussion}\label{sec:experiments}
Having outlined our approach to data collection, we fine-tune LLaVA-v1.5-13B on our augmented dataset to realize {\fontfamily{cmtt}\selectfont Space-LLaVA}, an open-source VLM adapted to extraterrestrial applications. {\fontfamily{cmtt}\selectfont Space-LLaVA} serves as a first proof of concept for a number of capabilities that make the use of FMs attractive in space robotics, e.g., serving as a foundation for \emph{specialization}, demonstrating \emph{emergent abilities} and mitigating \emph{catastrophic forgetting}. In our experiments, we demonstrate these capabilities by comparing the model's performance against SoTA FMs applied zero-shot to in-distribution and withheld training task types. Explicitly, we investigate 1) whether adaptation is necessary for SoTA VLMs in an extraterrestrial environment, and if so, to what degree components of the model must be adapted; 2) whether our augmented dataset instantiated as e.g., instruction-following, conversations enhances {\fontfamily{cmtt}\selectfont Space-LLaVA's} zero-shot performance on previously unseen task types; and 3) to what extent---if at all---fine-tuning stains LLaVA's generalist prior as measured by performance on standard VQA benchmarks.

\begin{figure}[b!]
    \centering
    \includegraphics[width=\linewidth]{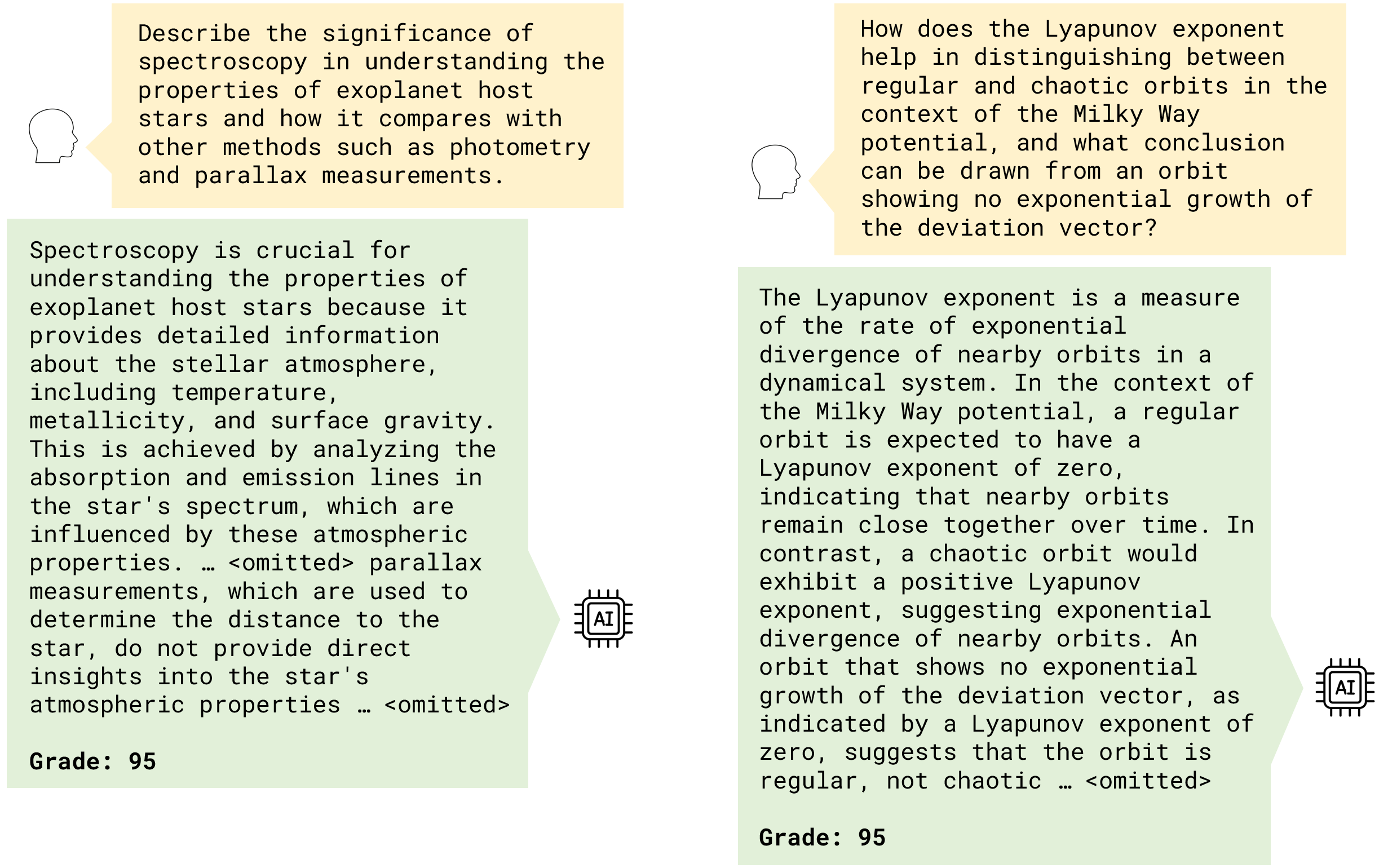}
    \caption{{\fontfamily{cmtt}\selectfont Space-LLaVA} responds to withheld queries from our {\fontfamily{cmtt}\selectfont SpaceScienceQA dataset}. Each response is scored on a scale of 0-100 points by GPT-4 for consistency with the ground-truth answer, while verbose sections of the response are omitted for brevity.}
    \label{fig:space-science-qa-generation}
\end{figure}

\subsection{Experimental Setup}\label{sec:experimental_setup}
Recall from \Cref{fig:spacellava} that the LLaVA model is comprised of three components: a Vision Encoder (VE), a Multi-Modal Adapter (MMA), and a Language Model (LM). In this work, we experiment with training three combinations of LLaVA's components: (1) training only the language model backbone; (2) training only the multi-modal adapter; (3) training the language model along with the multi-modal adapter together. We do not experiment with fine-tuning LLaVA's vision encoder as our synthetic, instruction-following dataset is not compatible with the contrastive objective typically used to train these encoders \cite{radford2021clip}. We also ablate adding 90\% and 20\% of the LLaVA-Instruct-150k \cite{liu2023visual} dataset to our training pipeline for a total of six training configurations. Importantly, in order to test our model's inference on unseen task types, akin to early research in instruction-tuning \cite{wei2022finetuned}, we hold-out the task of {\fontfamily{cmtt}\selectfont terrain comparison} from training to serve as a withheld evaluation task.

We train each configuration of the model with hyper-parameters, e.g., learning rate and weight decay, inspired by \cite{Marcu2023LingoQAVQ, pgvlm2024} and empirical results from our initial experimentation \cite{foutter2024adaptingfoundationmodelspacebased}. We train with an 80-20 train-validation split on the {\fontfamily{cmtt}\selectfont Space-LLaVA dataset} for 4 epochs with a batch size of 256 and learning rate of $3\times 10^{-5}$ on a compute infrastructure with 8x 80GB A100 GPUs. As introduced in \Cref{sec:problem_formulation}, we evaluate the strength of each training configuration against the zero-shot responses produced by SoTA VLMs, e.g., GPT-4o. That is, we measure performance by 1) judging the rate at which {\fontfamily{cmtt}\selectfont Space-LLaVA's} responses are preferable to a SoTA VLM and 2) scoring both responses on a numeric scale from 0 (worst) to 100 (exemplar) with further detail provided in \Cref{sec:appendix_pref_analysis}.

\subsection{Experimental Results}\label{sec:experimental_results}

\begin{figure}[t]
    \centering
    \includegraphics[width=0.9\linewidth]{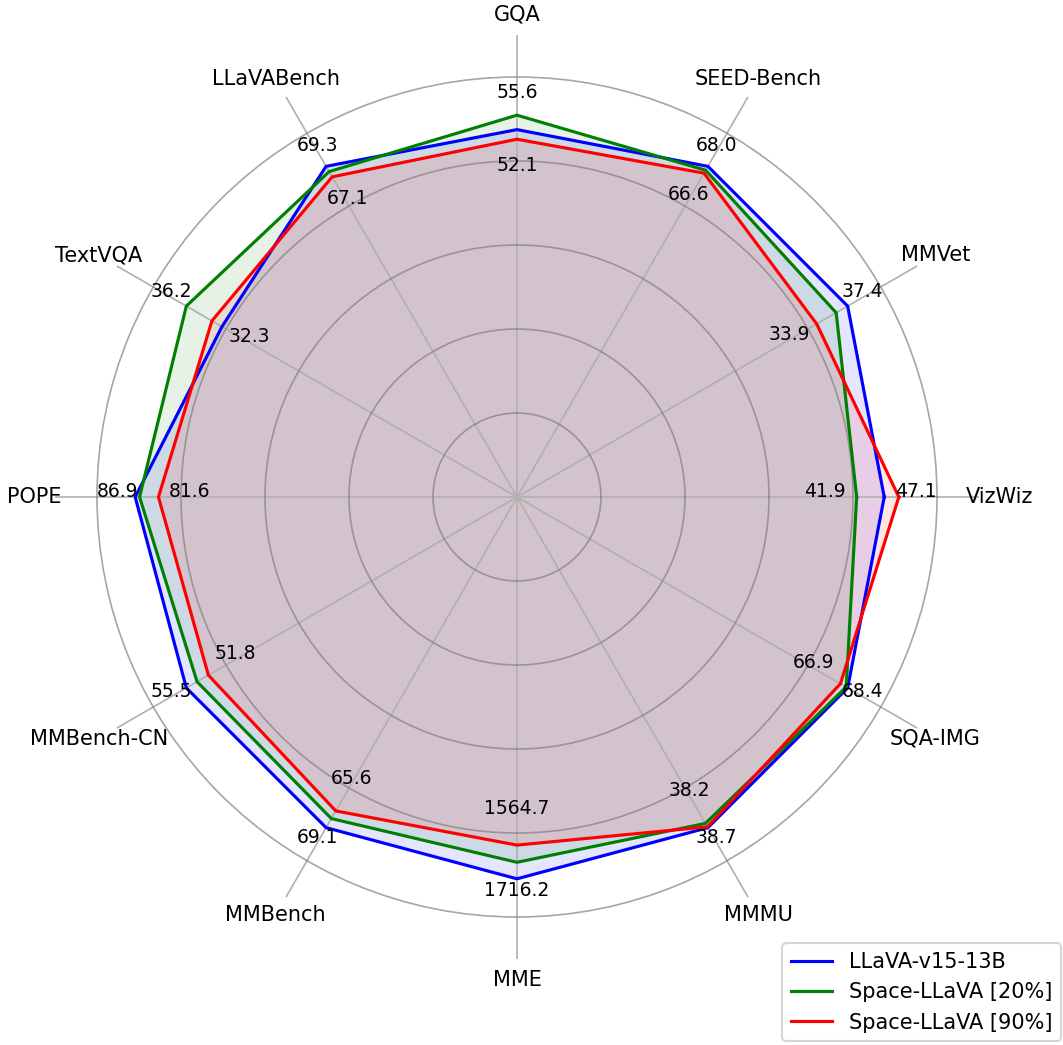}
    \caption{{\fontfamily{cmtt}\selectfont Space-LLaVA}---with a frozen vision encoder---co-trained at 20\% achieves comparable performance to base LLaVA on standard VQA benchmarks \cite{Hudson2019GQAAN, Li2023SEEDBenchBM, yu2024mm, Gurari2018VizWizGC, lu2022learn, yue2023mmmu, Fu2023MMEAC, MMBench, Li-hallucination-2023, singh2019towards, Liu2023ImprovedBW}. Data labels are provided along each axis for the best and worst scoring model. {\fontfamily{cmtt}\selectfont Space-LLaVA [20\%]} routinely scores within 5\% of base LLaVA, exceeding the base model in several instances. All evaluations are supported by an existing code-base \cite{duan2024vlmevalkit}.}
    \label{fig:forgetting} 
\end{figure}

The results of fine-tuning each configuration in comparison to zero-shot LLaVA and GPT-4o performing the task of {\fontfamily{cmtt}\selectfont terrain comparison} are presented in \Cref{results}. In \Cref{fig:space-llava-performance-evaluation}, we present the average numeric score assigned to responses from {\fontfamily{cmtt}\selectfont Space-LLaVA}, zero-shot LLaVA and GPT-4o evaluated on withheld observations from in-distribution task types, e.g., {\fontfamily{cmtt}\selectfont scene description}, and the withheld task of {\fontfamily{cmtt}\selectfont terrain comparison}. Further, we provide two high-quality demonstrations from {\fontfamily{cmtt}\selectfont Space-LLaVA} on our {\fontfamily{cmtt}\selectfont SpaceScienceQA dataset} in \Cref{fig:space-science-qa-generation} and characterize {\fontfamily{cmtt}\selectfont Space-LLaVA} against LLaVA-v1.5-13B on standard VQA benchmarks in \Cref{fig:forgetting}.

Based on the results in \Cref{results} and \Cref{fig:space-llava-performance-evaluation}, it is immediately apparent that SoTA VLMs out-of-the-box are ill-equipped to process the novel semantic features in an extraterrestrial environment. Indeed, in \Cref{fig:space-llava-performance-evaluation}, we find that fine-tuning significantly improves the quality of language annotations on in-distribution tasks such as {\fontfamily{cmtt}\selectfont scene description} and {\fontfamily{cmtt}\selectfont group terrain localization} by 12.1\% and 34.4\%, respectively, in comparison to GPT-4o---the strongest zero-shot model. Our results in \Cref{fig:space-llava-performance-evaluation}, corroborated by \Cref{results}, show that our dataset enhances the quality of {\fontfamily{cmtt}\selectfont Space-LLaVA's} responses to previously withheld extraterrestrial-based task types---demonstrating a 19.3\% and 40.8\% performance increase relative to GPT-4o and base LLaVA, respectively. These evaluations showcase {\fontfamily{cmtt}\selectfont Space-LLaVA} as a first proof of concept that FMs can be fine-tuned in space robotics to 1) achieve SoTA performance on \emph{specialized}, offline tasks where a zero-shot FM may otherwise be necessary due to e.g., the budget or personnel constraints in managing satellite mega-constellations, and 2) proficiently answer unseen task types as an \emph{emergent ability}, offering promise for robots to autonomously solve unanticipated problems in a new environment.

Interestingly, \Cref{results} demonstrates that fine-tuning the multi-modal adapter alone is insufficient to achieve responses that are preferable to GPT-4o or distinguishable in comparison to base LLaVA; however, training the language model and multi-modal adapter in concert significantly outperforms SoTA VLMs providing the largest benefit for the sensory reasoning tasks captured by our dataset. This observation is likely because the multi-modal adapter, whose task is to project vision embeddings into a shared language embedding space, merely acts as a custodian for the semantic content in a vision embedding, potentially accentuating task-relevant semantics \cite{pgvlm2024}, but not to introduce features that were previously absent. Whereas if we fine-tune the language model, then we can directly tune the FMs reasoning apparatus and learn to associate new semantic concepts, e.g., the signature characteristics of ``bedrock" on Mars, with the features already extracted by the vision encoder. Hence, we posit that fine-tuning the language model significantly improves adaptation given that our dataset requires VLMs to perform \emph{fine-grained semantic reasoning} in extraterrestrial applications, whereas existing work, e.g., the VLM from \cite{pgvlm2024}, which only trains InstructBLIP's \cite{dai2023instructblip} equivalent to the multi-modal adapter, fine-tunes a VLM to perform \emph{image classification}. 

Nonetheless, \Cref{fig:space-llava-performance-evaluation} paints an incomplete picture. Recall, FMs offer promise to address the core challenges in space robotics as these models empirically demonstrate broad, semantic understanding \cite{sinha2024realtime, Elhafsi2023SemanticAD, pgvlm2024} and commonsense reasoning \cite{wei2022cot, wei2022finetuned, Kojima2022LargeLM}. However, straightforward fine-tuning of these general-purpose models to a narrow application---while effective in-distribution---has been shown to induce catastrophic forgetting \cite{zhai2023investigating, model_tailor}, severely limiting an FMs broad utility to the community. Co-training \cite{choi2024autoft, Huang2024MitigatingCF} is currently a popular method to safeguard against forgetting, generally characterized by balancing the fine-tuning dataset with a fraction of the data originally used in pre-training; akin to \cite{anonymous2024aha}, we choose to train {\fontfamily{cmtt}\selectfont Space-LLaVA} on our augmented dataset and the LLaVA-Instruct-150k dataset together to retain LLaVA's generalist prior during adaptation. \Cref{fig:forgetting} validates that our training regime preserves LLaVA's foundational skills, e.g., semantic understanding and basic visual reasoning, which compose the model's generalist prior, as {\fontfamily{cmtt}\selectfont Space-LLaVA 20\%} routinely matches, or even exceeds, the base LLaVA model on standard VQA benchmarks. Somewhat surprisingly, {\fontfamily{cmtt}\selectfont Space-LLaVA 90\%} routinely performs \emph{worse} than the same architecture with less data, which we hypothesize is due to the total number of gradient steps in training. That is, for a fixed number of epochs and batch size, {\fontfamily{cmtt}\selectfont Space-LLaVA 90\%} requires nearly \emph{double} as many gradient steps as the {\fontfamily{cmtt}\selectfont Space-LLaVA 20\%} model; hence, {\fontfamily{cmtt}\selectfont Space-LLaVA 20\%} preserves the base LLaVA model through inherently less fine-tuning and over-fitting.

Finally, we see that configurations of {\fontfamily{cmtt}\selectfont Space-LLaVA} with equivalent fine-tuned parameters exhibit comparable performance on {\fontfamily{cmtt}\selectfont terrain comparison} when trained on 20\% and 90\% of LLaVA-Instruct-150k. After all, we choose to co-train as a measure to mitigate catastrophic forgetting, i.e., LLaVA-Instruct-150k is an entirely terrestrial dataset, in which case \Cref{results} follows intuition as one may expect {\fontfamily{cmtt}\selectfont Space-LLaVA's} performance on {\fontfamily{cmtt}\selectfont terrain comparison} to be derived directly from our synthetic dataset.

\section{An Out-of-Distribution Evaluation: Orbital Space} \label{sec:orbit}

The previous experiments, as presented in \Cref{fig:forgetting}, validate {\fontfamily{cmtt}\selectfont Space-LLaVA} as a general-purpose tool for basic perception, commonsense reasoning, and cognition. However, these evaluations remain incomplete without an investigation into {\fontfamily{cmtt}\selectfont Space-LLaVA’s} sensitivity to out-of-domain (OOD) scenarios—specifically, withheld extraterrestrial task types and application domains such as on-orbit imagery. More broadly, the expertise required to score well on VQA benchmarks is primarily composed of pre-training, or pre-training-adjacent, skills, and since no measure is taken to stabilize our model on withheld extraterrestrial applications, there remains an open concern whether {\fontfamily{cmtt}\selectfont Space-LLaVA} is trustworthy in ``far-OOD" application domains. Therefore, for completeness, we develop an OOD evaluation grounded in the visually sparse and distinct domain of on-orbit imagery, probing {\fontfamily{cmtt}\selectfont Space-LLaVA's} generalist capabilities in a withheld space robotics application. In this section we discuss motivating examples for the use of FMs in orbital space, outline our methodology to curate a ``far-OOD" evaluation dataset, and benchmark {\fontfamily{cmtt}\selectfont Space-LLaVA} against the base LLaVA model.

\subsection{Objective}

Satellite mission operators often need to filter large amounts of on-orbit imagery in order to select data which possess desirable characteristics. Consider the following motivating examples, in which, for instance, a satellite operator may wish to collect: 1) close-range camera imagery of specific components, e.g., the solar panel, on a target satellite in an ISAM scenario; 2) star tracker images containing transient Resident Space Objects (RSOs) to support a secondary SSA objective; 3) in-focus and well-lit imagery of an RSO target to support continual learning \cite{Wang2023ACS} for a learning-based satellite pose estimation network \cite{park_online_2024}.

Traditionally, these objectives require a human-in-the-loop for domain-specific knowledge and open-ended visual reasoning; however, the operator may need to process thousands of images, which does not scale with the continued growth of satellite operations. Algorithms have been developed for specific objectives, e.g., image classification \cite{rawat_classify_2017}, OOD detection \cite{foutter2023selfsupervised} and RSO streak detection \cite{virtanen_streak_2016}, but these algorithms are narrowly-scoped by nature, require considerable development and may require extensive re-training for adapting to evolving mission or user requirements. 

Therefore, we investigate whether {\fontfamily{cmtt}\selectfont Space-LLaVA's} open-ended, visual reasoning can be used to address the three aforementioned objectives, probing the model's performance in three OOD scenarios relative to the base LLaVA model. To the best of our knowledge, current literature has not explored the ability of VLMs to filter flight image data in support of on-orbit operations for either open-source (e.g., LLaVA) or closed-source (e.g., GPT-4o) models.

\begin{figure}[t]
    \centering
    \includegraphics[width = \textwidth]{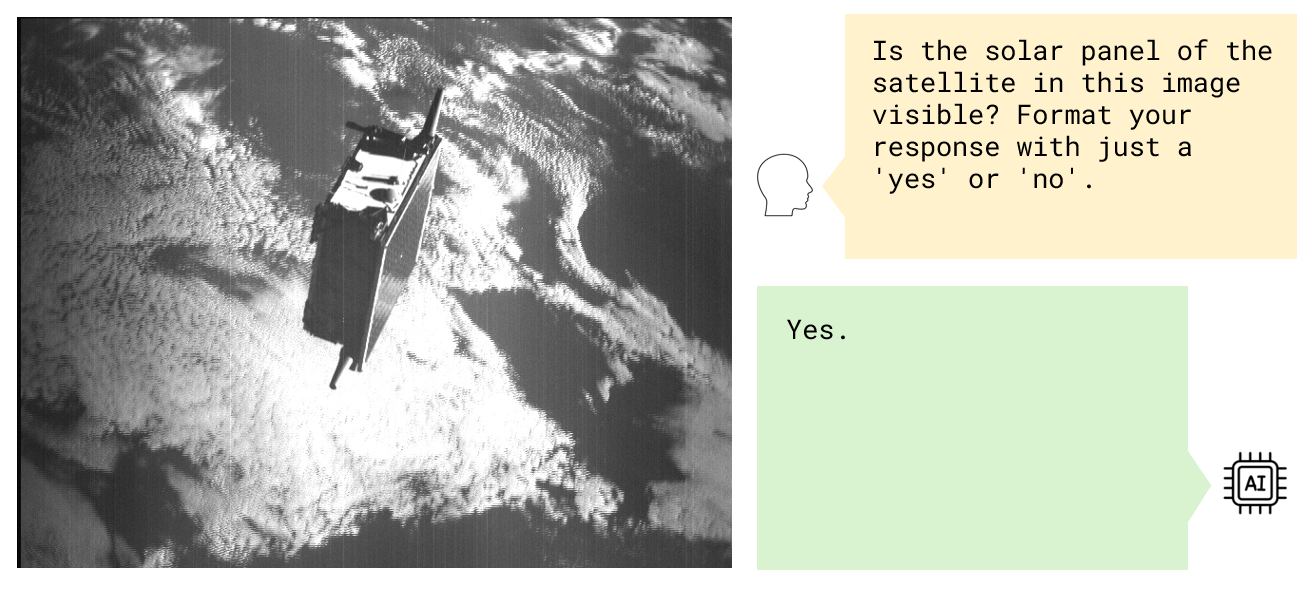}
    \caption{An example \textit{Mango} image and corresponding FM prompt.}
    \label{fig:prisma_example}
\end{figure}

\subsection{Experimental Setup}

We collect a test dataset of several thousand flight images from three existing satellite imagery databases to support this proof of concept. We prefer to leverage in-flight data due to the nuanced visual characteristics which are often challenging to achieve in simulation \cite{park2022speedp}.

Our evaluation dataset encompasses on-orbit imagery from the \emph{Mango} spacecraft of the PRISMA mission \cite{damico_argon_2013}, the \emph{Starling} SV2 CubeSat \cite{kruger_flight_2024} and the \emph{ExoRomper} payload of the Slingshot program \cite{weiher_exoromper_2022}. 
In accordance with the three aforementioned examples, we curate questions to filter on-orbit imagery as follows: 1) in support of the ISAM objective, we ask ``is the solar panel of the target satellite visible, to facilitate inspection?" on the \emph{Mango} dataset; 2) in support of the second SSA objective, we ask ``does the star tracker image contain any unusual features atypical of a star tracker image?" on the \emph{Starling} dataset; and 3) in support of the continual learning objective, we ask ``is the target satellite in plain view, not occluded, and well-lit?" on the \emph{ExoRomper} dataset.
%
We provide a representative example of this evaluation for the ISAM objective in  \Cref{fig:prisma_example} and image-prompt examples for all other applications in \Cref{sec:appendix_orbital_space}. 

We evaluate a VLM's performance towards each of these objectives by measuring the model's binary classification accuracy, e.g., whether the solar panel is in view, against ground-truth labels determined by human operators. The specific criteria used in curating the orbital space dataset is outlined in \Cref{sec:appendix_orbital_space}. Specifically, we evaluate LLaVA-v1.5-13B \cite{Liu2023ImprovedBW} and {\fontfamily{cmtt}\selectfont Space-LLaVA} zero-shot on the orbital space dataset to examine the impact of fine-tuning on the model's ability to generalize out-of-domain.

\subsection{Results and Discussion}

In our experiments we benchmark two {\fontfamily{cmtt}\selectfont Space-LLaVA} configurations---with a frozen vision encoder---fine-tuned with 20\% and 90\% of the LLaVA-Instruct-150k dataset. We characterize the performance of these models on our orbital dataset through binary classification statistics, namely, the True Positive Rate (TPR), False Positive Rate (FPR), and F-measure.

\begin{table}[b]
\begin{tabular}{ccccc}
\toprule
 & & \multicolumn{3}{c}{Evaluation Dataset} \\
\cmidrule{3-5}
Model & Metric & PRISMA & Starling & Exo. \\
\midrule
\multirow{2}{*}{\begin{tabular}[c]{@{}l@{}}LLaVA-\\v1.5-13B\end{tabular}} & TPR & 60.29\% & 79.74\% & 45.43\% \\

& FPR &  19.95\% & 60.74\% & 26.58\% \\

& $F_1$ &   \textbf{0.314}    &   \underline{0.610}    &   \underline{0.433}    \\ \midrule

\multirow{2}{*}{\begin{tabular}[c]{@{}l@{}}{\fontfamily{cmtt}\selectfont Space-}\\ {\fontfamily{cmtt}\selectfont LLaVA 20\%}\end{tabular}} & TPR & 52.94\% &  93.16\% & 68.70\% \\
& FPR &  31.23\% & 80.47\% & 60.48\% \\

& $F_1$ & 0.211 & \textbf{0.618} & \textbf{0.436} \\ \midrule
\multirow{2}{*}{\begin{tabular}[c]{@{}l@{}}{\fontfamily{cmtt}\selectfont Space-}\\ {\fontfamily{cmtt}\selectfont LLaVA 90\%}\end{tabular}} & TPR & 57.35\% & 69.47\% & 42.66\% \\
& FPR & 33.33\% & 45.51\% & 32.30\% \\
& $F_1$ &   \underline{0.216}    &   0.602    &   0.386      \\
\bottomrule
\end{tabular}
\caption{Zero-shot performance on data filtering between base LLaVA and {\fontfamily{cmtt}\selectfont Space-LLaVA} trained with 20\% and 90\% of LLaVA-Instruct-150k. We report measure through binary classification statistics, e.g., True Positive Rate (TPR), False Positive Rate (FPR) and F-measure. The strongest and second-best F-measure is bold and underlined, respectively, for each dataset.}
\label{table:satellite_zeroshot}
\end{table}

The F-measure in \Cref{table:satellite_zeroshot} reveals that both {\fontfamily{cmtt}\selectfont Space-LLaVA 20\%} and {\fontfamily{cmtt}\selectfont Space-LLaVA 90\%} demonstrate similar performance in comparison to base LLaVA. This result is particularly notable given the significant domain gap and the sparse visual characteristics of orbital space data relative to the planetary imagery used in fine-tuning. We highlight that while these open-source models do not demonstrate flight-ready performance, 
the consistency between models within each dataset suggests that the poor performance can be attributed to the base LLaVA model's skill base. This implies that fine-tuning on orbital imagery could significantly improve performance on our benchmark, potentially even when fine-tuned on different in-orbit tasks, as shown in \Cref{sec:experiments}. Therefore, the findings of this section demonstrate a need to explore further avenues for deploying VLMs in orbital contexts and their promise for automating spacecraft operations.

\section{Agency: Lunar Simulation}

\begin{figure*}[ht]
    \centering
    \includegraphics[width = 0.9\linewidth]{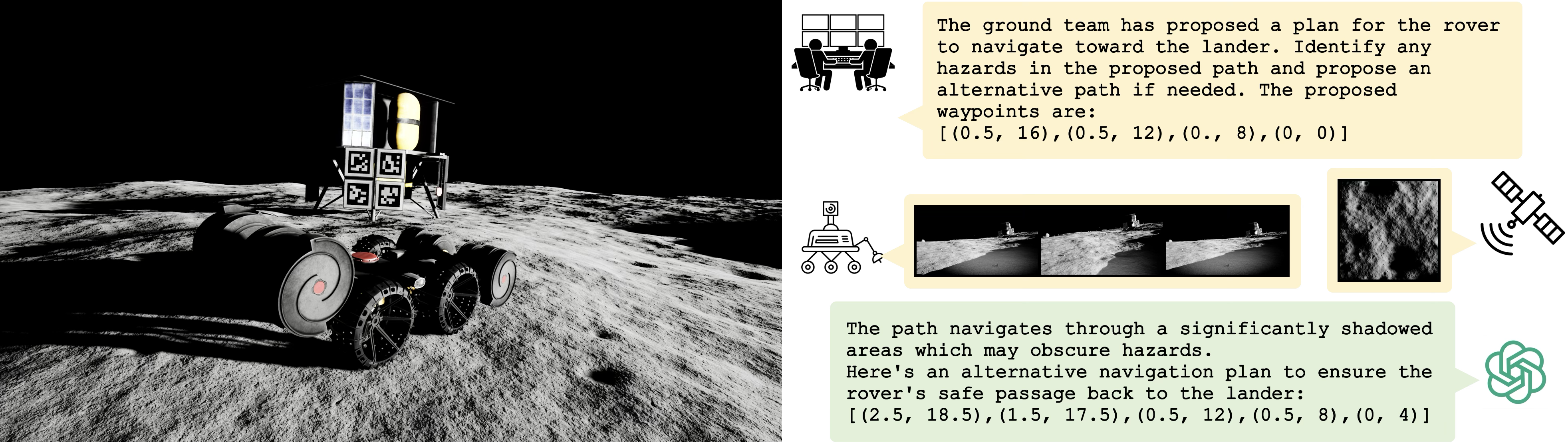}
    \caption{The lunar mobility scenario. (Left) A lunar rover and lander are situated in a virtual lunar environment. The rover, equipped with multiple onboard cameras, must navigate from its starting position to the lander, guided by a candidate path plan, such as one provided by a hypothetical ground team. (Right) The FM serves as a high-level path planner and runtime monitor, evaluating the safety and feasibility of the proposed plan using both onboard and off-board images, and, if necessary, generating an alternative path.}
    \label{fig:lunarsim}
\end{figure*}

Finally, we explore the potential of integrating FMs into modular autonomy stacks, a common architecture in space robotics platforms \cite{GripEtAl2018}. 
Specifically, we depart from the evaluation of {\fontfamily{cmtt}\selectfont Space-LLaVA} and focus on broadly demonstrating the use of an FM as a high-level path planner and runtime monitor for a lunar rover. 
The FM’s role is to serve as a flexible, multi-modal reasoning module that interfaces with pre-existing components within the rover’s autonomous mobility stack. 

In what follows, we first introduce the lunar mobility scenario, highlight the role of the FM within it, and discuss its broader significance for the space robotics community. Then, we describe the experimental setup and present key empirical results and insights.

\textbf{Lunar Mobility Application and Broader Interest:} The scenario for this experiment, illustrated in \Cref{fig:lunarsim}, can be summarized as follows:

\textit{Context:} A rover and a lander are positioned in an under-explored region of the lunar surface. 
The area of interest is represented by a coordinate system, where each $(x, y)$ denotes a unique location. 
The rover, equipped with multiple onboard cameras, receives from the ground team a candidate path to traverse from its current position $(x_R, y_R)$ to the lander’s position $(x_L, y_L)$, e.g., for recharging or maintenance. 
The path is defined by $M$ waypoints, $W = [w_1, \ldots, w_M]$, where $w_i = (x_i, y_i)$.  

\textit{Goal:} The FM is tasked with (1) evaluating the safety and feasibility of the proposed path, and (2) suggesting an alternative path if the initial plan is deemed unsafe or suboptimal. 
This requires balancing safety with mission-related objectives.

\textit{Inputs (Image, Text):} Image---RGB images from the rover’s onboard cameras, along with a top-down view of the lunar surface overlaid with a grid for geometric reference; Text---A task description and the list of proposed waypoints $W$.  

\textit{Output (Text):} The FM generates a textual analysis of the proposed path, including safety and feasibility evaluations, and, if needed, an alternative set of waypoints.  

\textit{Interface:} The FM’s output is parsed and used as input to a low-level motion planning module, which computes the actuator commands required to navigate the rover to the high-level waypoints.  

More broadly, this lunar mobility application represents an instantiation of a broader framework for integrating FMs into space robotics operations, one where FMs can enable: (1) \emph{Multi-modal input processing}, combining data from onboard and off-board sensors with mission specifications articulated in natural language, (2) \emph{Automated reasoning capabilities}, analyzing inputs to generate actionable insights aligned with mission objectives, and (3) \emph{Reconfigurable interfaces}, seamlessly integrating with diverse lower-level components to support modularity and adaptability across various tasks and mission contexts, with outputs conveyed in natural language.
Crucially, by abstracting high-level reasoning from low-level execution, FMs provide a powerful tool for enhancing autonomy and operational efficiency in space exploration, while maintaining the modularity of traditional approaches.

\subsection{Experimental Setup}
We conducted autonomous lunar mobility experiments using a simulator platform that provides realistic 3D environments for testing and validating autonomous rover operations on the lunar surface\footnote{At the time of writing, the simulator utilized for our experiments remains inaccessible to the public. Consequently, we refrain from providing any details that could disclose the simulator’s functionalities and focus exclusively on presenting the analyses and insights derived from our experimental results.}.
The goal of these experiments is to demonstrate that the FM-based monitor can effectively identify hazardous path plans and propose alternative solutions that balance rover safety and mission success.
We designed three experimental scenarios within the constraints of the simulation environment:
\begin{enumerate}[wide = 0pt, labelwidth = 1.3333em, labelsep = 0.3333em, leftmargin = \dimexpr\labelwidth + \labelsep\relax ]
    \item Low-visibility regions: Shadows caused by low solar angles on the lunar surface obscure potential hazards, such as uneven terrain or concealed obstacles, making safe navigation more challenging.
    \item Large obstacles and structural threats: Large rocks, modeled as immovable objects, pose a risk of causing irreparable damage to the rover.
    \item Hazardous terrain and uneven regolith: Simulated as a deformable surface mimicking regolith properties, certain regions of the lunar surface feature steep slopes and extreme irregularities, presenting substantial risks to rover stability and mobility.
\end{enumerate}

To simulate the autonomous rover's behavior, we developed a goal-reaching low-level policy that navigates the rover sequentially to each waypoint in the provided path until it reaches the final destination.
For perception, we used a combination of three forward-facing RGB cameras (i.e., Front Left, Front, Front Right) fixed on the rover's body and a single top-down image of the operational area provided by e.g., a Lunar Orbiter spacecraft to aid surface operations.

To detect hazards in the proposed path and compute the waypoints for an alternative route to the lander, we generated prompts based on the rover's observations and top-down views, which were then used as input to GPT-4o. 
The prompt---used in tandem with the four camera images---was designed to elicit a chain-of-thought reasoning style to assess whether any elements in the scene could pose a hazard to the rover’s safe operation along the path suggested by the ground team.
An illustration of this process is illustrated in \Cref{sec:appendix_lunar_sim}. 

\subsection{Experimental Results}
\begin{figure}[ht!]
    \centering
    \begin{subfigure}{\linewidth}
        \centering
        \includegraphics[width=0.95\linewidth]{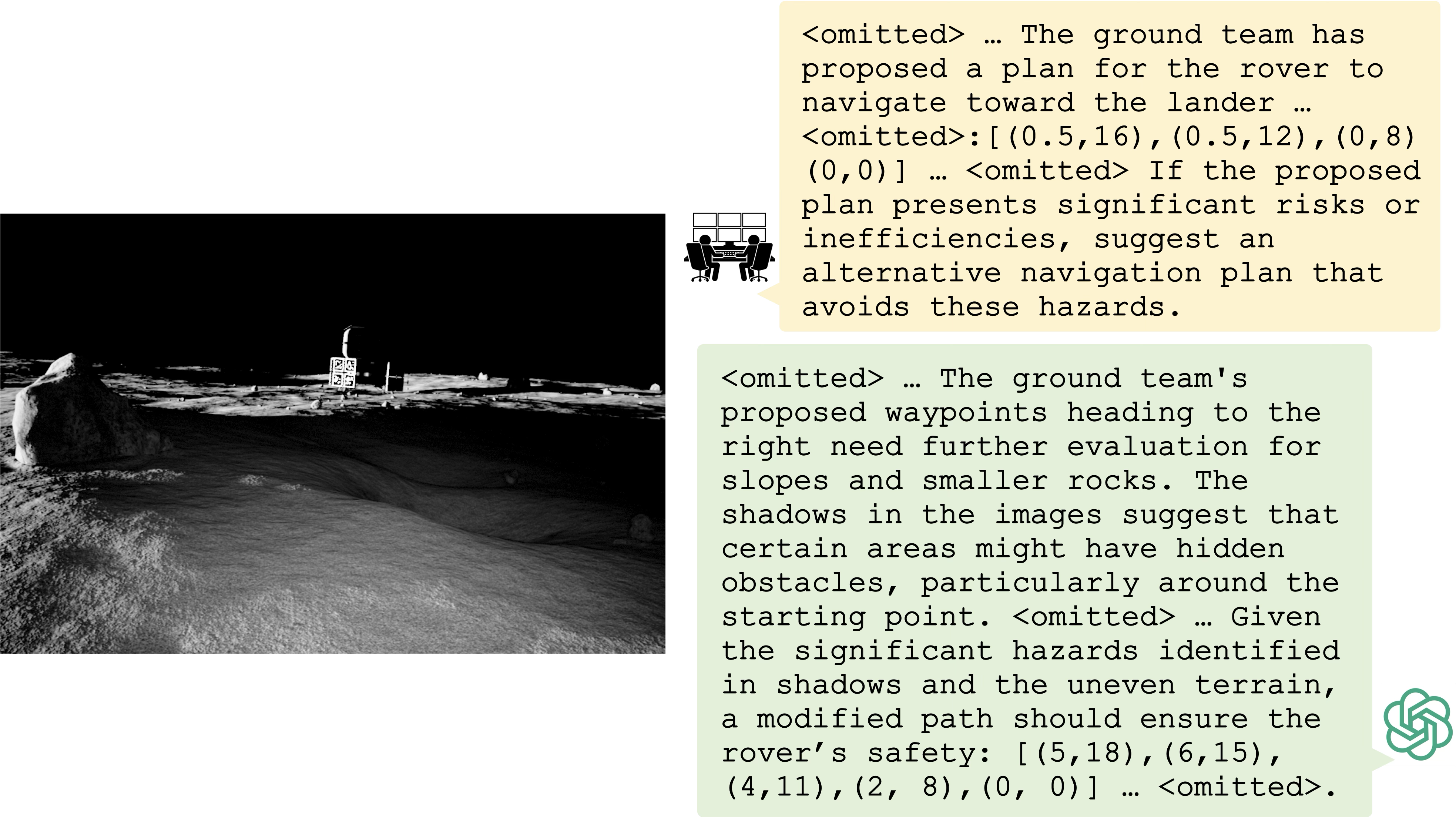}
        \caption{Example generation. GPT-4o evaluates the safety and feasibility of the proposed path based on onboard and off-board images, ultimately suggesting an alternative path that avoids major safety hazards.}
        \label{fig-lun-sim:gpt4-generation}
        \vspace{1.25em}
    \end{subfigure}
    \vspace{0.25em}
    \begin{subfigure}{\linewidth}
        \centering
        \includegraphics[width=0.95\linewidth]{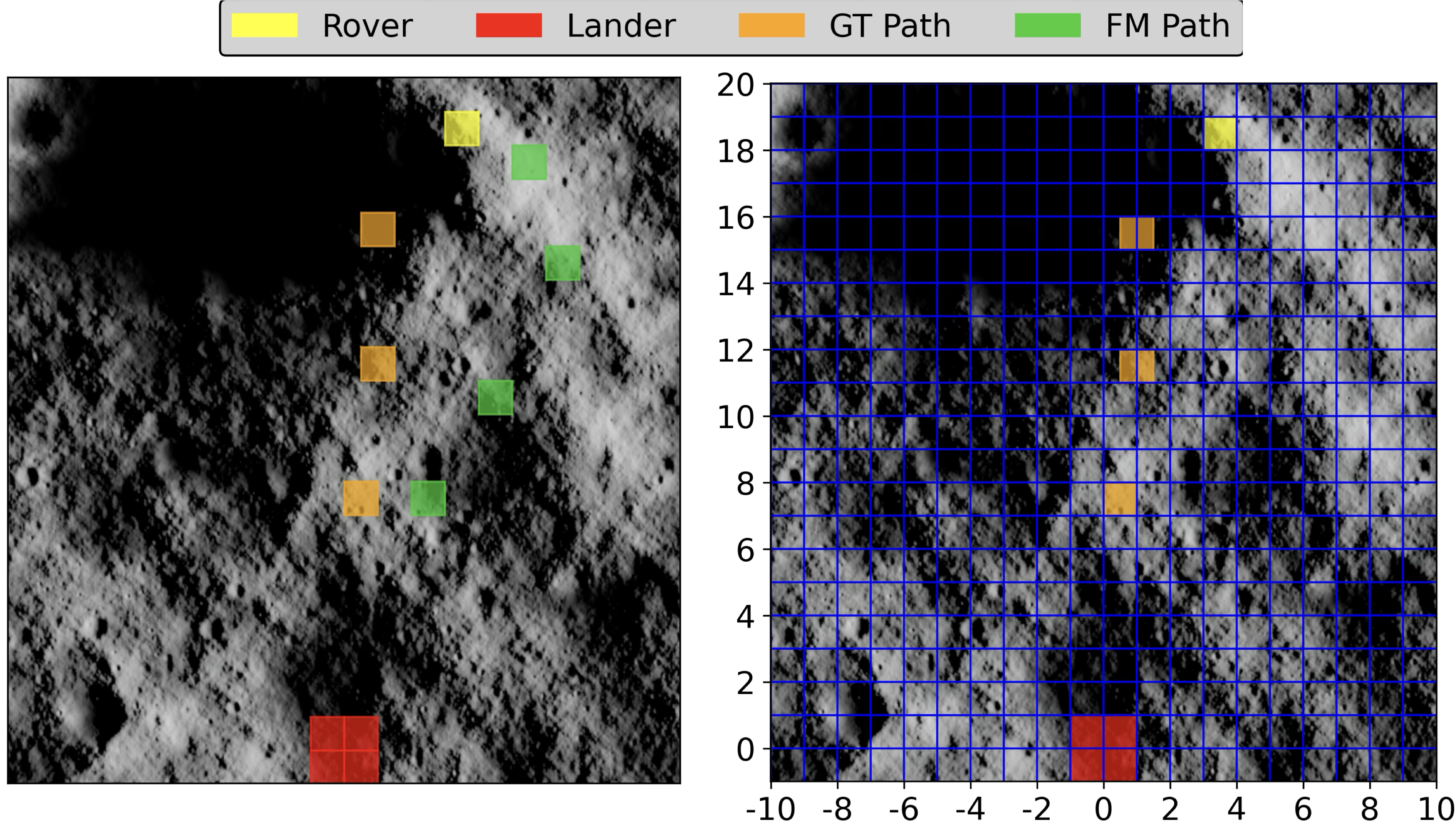}
        \caption{(Left) A top-down visualization showing the ground team’s (i.e., GT) proposed path (orange) alongside GPT-4o’s suggested alternative path (green). (Right) The top-down input image provided to GPT-4o, featuring a superimposed spatial grid to support the FM’s geometric reasoning. Video available at: \url{https://www.youtube.com/watch?v=gvWdJLQXtbU}}
        \label{fig-lun-sim:top-down-generation}
    \end{subfigure}
    \caption{Hazardous terrain and uneven regolith. (a) Example generation. (b) Scenario visualizations.}
    \label{fig-lun-sim:generation}
\end{figure}

We evaluate the performance of the FM as a general-purpose path planner and runtime monitor by qualitatively inspecting its hazard detections and assessing the feasibility of the proposed plan when executing it in closed-loop within the lunar simulator. 
The results of this analysis across the three designed scenarios are presented in \Cref{fig-lun-sim:generation} and \Cref{sec:appendix_lunar_sim}\footnote{Videos of the closed-loop execution are available at  \\ \url{https://www.youtube.com/playlist?list=PL8-2mtIlFIJpc-RZNQk4svyqd2WtXwMaT}.}.

We observe that SoTA VLMs exhibit promising performance across the considered scenarios, with the model consistently correlating geometric information—such as the coordinate values of the ground team’s proposed path—with visual data from both on- and off-board camera images. 
Moreover, the VLM successfully detects the correct hazards in all scenarios, such as large rocks, uneven terrain, and low-visibility regions, demonstrating its ability for non-trivial, task-oriented reasoning.
This detailed scene understanding is further validated by the VLM’s ability to generate alternative path plans that align with its interpretation of the scene. 
Crucially, when executed in closed-loop, these plans guide the rover to its destination while avoiding the detected hazards.

While a diverse set of experiments are described in \Cref{sec:appendix_lunar_sim}, \Cref{fig-lun-sim:generation} highlights a representative example of the VLM's generation.
Specifically, in the scenario depicted in \Cref{fig-lun-sim:gpt4-generation}, the rover is positioned near a rock (on the left side of the image) and a steep, low-visibility region (on the right side of the image).
The path proposed by the ground team---depicted as the orange cells in the top-down view from \Cref{fig-lun-sim:top-down-generation}---leads directly into the uneven terrain on the right, presenting a potential safety hazard.
The summarized generation in \Cref{fig-lun-sim:gpt4-generation} illustrates how the VLM successfully detects the hazards in the scene and correlates them with the ground team's proposed path. Ultimately, the VLM suggests an alternative path, shown as the green cells in the top-down view from \Cref{fig-lun-sim:top-down-generation}, that avoids both the rock and the uneven terrain by taking a sharp left turn before proceeding toward the lander.

\section{Conclusion}\label{sec:conclusion}

In this paper, we highlight that core challenges in the future of space robotics motivate the development of a Foundation Model (FM) for the space robotics community. As a first step towards a \emph{space foundation model}, we augment three extraterrestrial databases with high-quality, GPT-assisted language annotations and adapt a pre-trained LLaVA checkpoint to the fine-grained semantic features in this planetary dataset, introducing {\fontfamily{cmtt}\selectfont Space-LLaVA}. We demonstrate that 1) existing Vision-Language Models (VLMs) are deficient visual reasoners in space-based applications, and 2) our visual instruction-tuning dataset showcases FMs as a foundation for \emph{specialization} and enhances {\fontfamily{cmtt}\selectfont Space-LLaVA's} zero-shot performance on unseen inference tasks as an \emph{emergent ability} in comparison to state-of-the-art VLMs. Subsequently, we validate that our training regime preserves {\fontfamily{cmtt}\selectfont Space-LLaVA's} generalist prior through an evaluation on standard VQA benchmarks and out-of-domain on satellite imagery. Future work in the development of a \emph{space foundation model} will incorporate: 1) collecting a sufficiently large and diverse space dataset, e.g., remote sensing data, spaceflight simulations, and space object catalogues, for space-related tasks and 2) developing data encoders to process the diverse modalities (LiDAR, GPS, etc) inherent to these data in order to create a meaningful representation for decision making.

\acknowledgments
This work is supported by Blue Origin (SPO \#299266) and Redwire (SPO \#299330) as Associate Members and Co-Founders of the Stanford's Center for AEroSpace Autonomy Research. This article solely reflects the opinions and conclusions of its authors and not any Blue Origin or Redwire entity. We also acknowledge OHB Sweden, the NASA Ames Research Center, and The Aerospace Corporation for providing PRISMA, Starling, and ExoRomper imagery, respectively.

\section*{Author Contributions}
\textbf{Matthew Foutter} initiated and lead the project, devised the {\fontfamily{cmtt}\selectfont Space-LLaVA dataset}, fine-tuned {\fontfamily{cmtt}\selectfont Space-LLaVA}, scoped and conducted the associated evaluations. \textbf{Daniele Gammelli} advised the project, consulted on the {\fontfamily{cmtt}\selectfont SpaceScienceQA dataset} and lead integration of GPT-4o into a lunar simulation environment. \textbf{Justin Kruger} lead and contributed to curating the orbital space dataset. \textbf{Ethan Foss} implemented and conducted the orbital space evaluation. \textbf{Praneet Bhoj} contributed to early investigations in the {\fontfamily{cmtt}\selectfont SpaceScienceQA dataset}. \textbf{Tommaso Guffanti} advised and contributed to curating the orbital space dataset. \textbf{Simone D'Amico} advised the project. \textbf{Marco Pavone} was the primary advisor for the project. The manuscript was jointly written by Matthew, Daniele, Justin, Ethan and Tommaso. All authors reviewed and revised the manuscript.

\newcommand{\newblock}{}
\bibliographystyle{IEEEtran}
\bibliography{references}

\thebiography
\begin{biographywithpic}{Matthew Foutter}{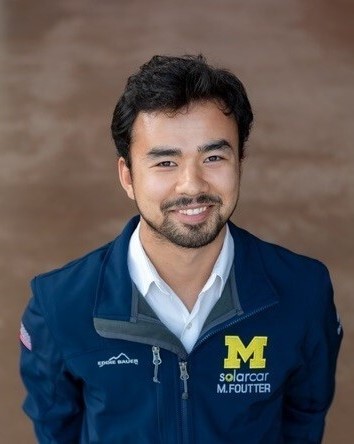}
Matt is a PhD candidate in the Autonomous Systems Lab at Stanford University. Matt’s research interests lie at the intersection of machine learning and robotics with the goal of enabling an autonomous robot to safely navigate in an unfamiliar environment. Specifically, he is interested in developing methodologies that monitor a robot’s operation at deployment - preemptively catching failure modes and enacting safety preserving actions.

Prior to joining Stanford, Matt graduated summa cum laude from the University of Michigan, Ann Arbor, with a B.S.E in Aerospace Engineering and minor in Computer Science. Also, as an undergraduate, he raced a solar car 1,800 miles across Australia and interned at SpaceX and MIT Lincoln Lab.
\end{biographywithpic}

\begin{biographywithpic}{Daniele Gammelli}{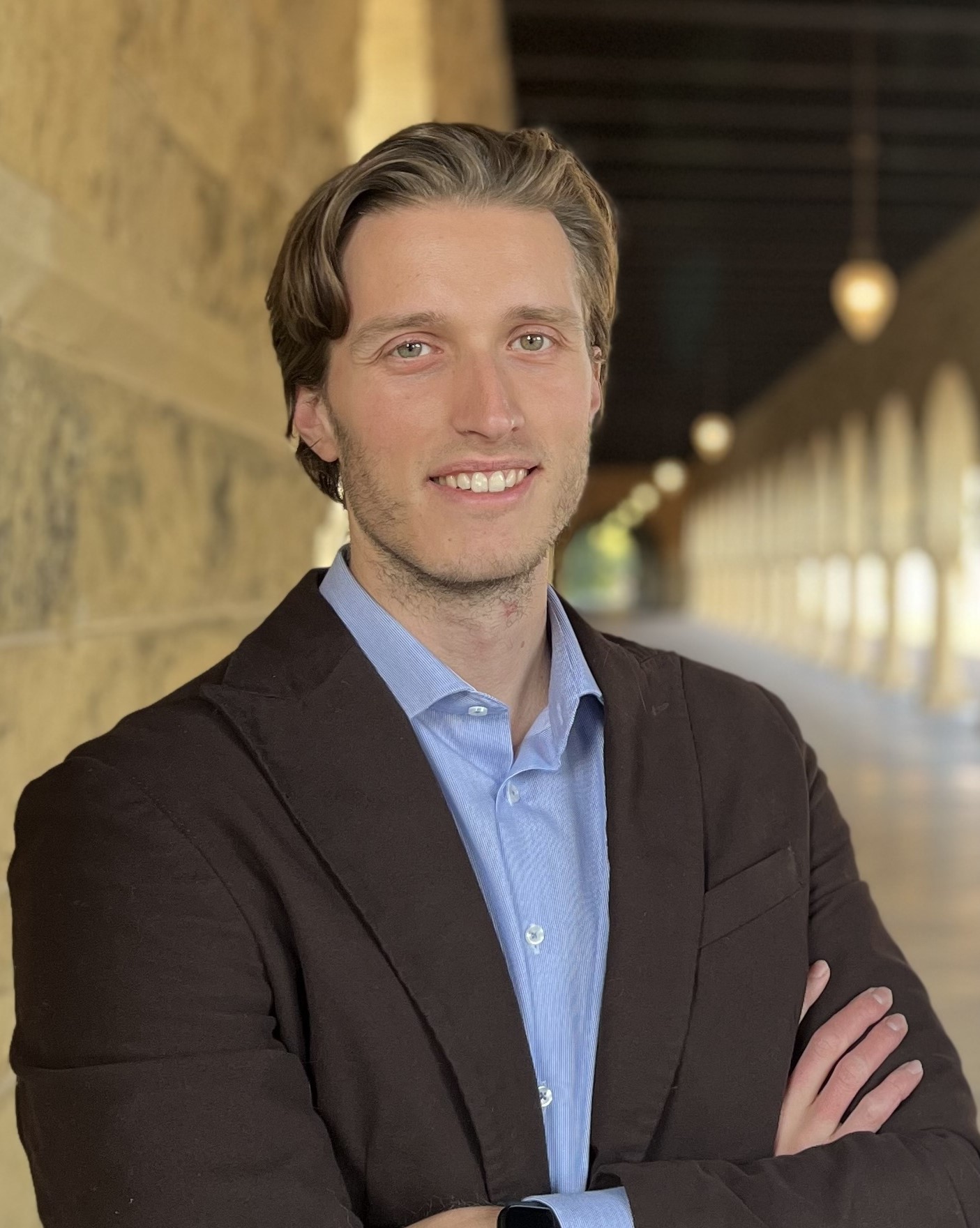}
Dr. Daniele Gammelli is a Postdoctoral Scholar in the Autonomous Systems Lab at Stanford University. He received the Ph.D. in Machine Learning and Mathematical Optimization at the Department of Technology, Management and Economics at the
Technical University of Denmark. Dr.
Gammelli’s research focuses on developing learning-based solutions that enable the deployment of future autonomous systems in complex
environments, with an emphasis on large-scale robotic networks, aerospace systems, and future mobility systems. During his doctorate and postdoctorate career, Dr. Gammelli
has been making research contributions in fundamental AI
research, robotics, and its applications to network optimization and mobility systems. His research interests include deep
reinforcement learning, generative models, graph neural networks, Bayesian statistics, and control techniques leveraging
these tools.
\end{biographywithpic}

\begin{biographywithpic}{Justin Kruger}{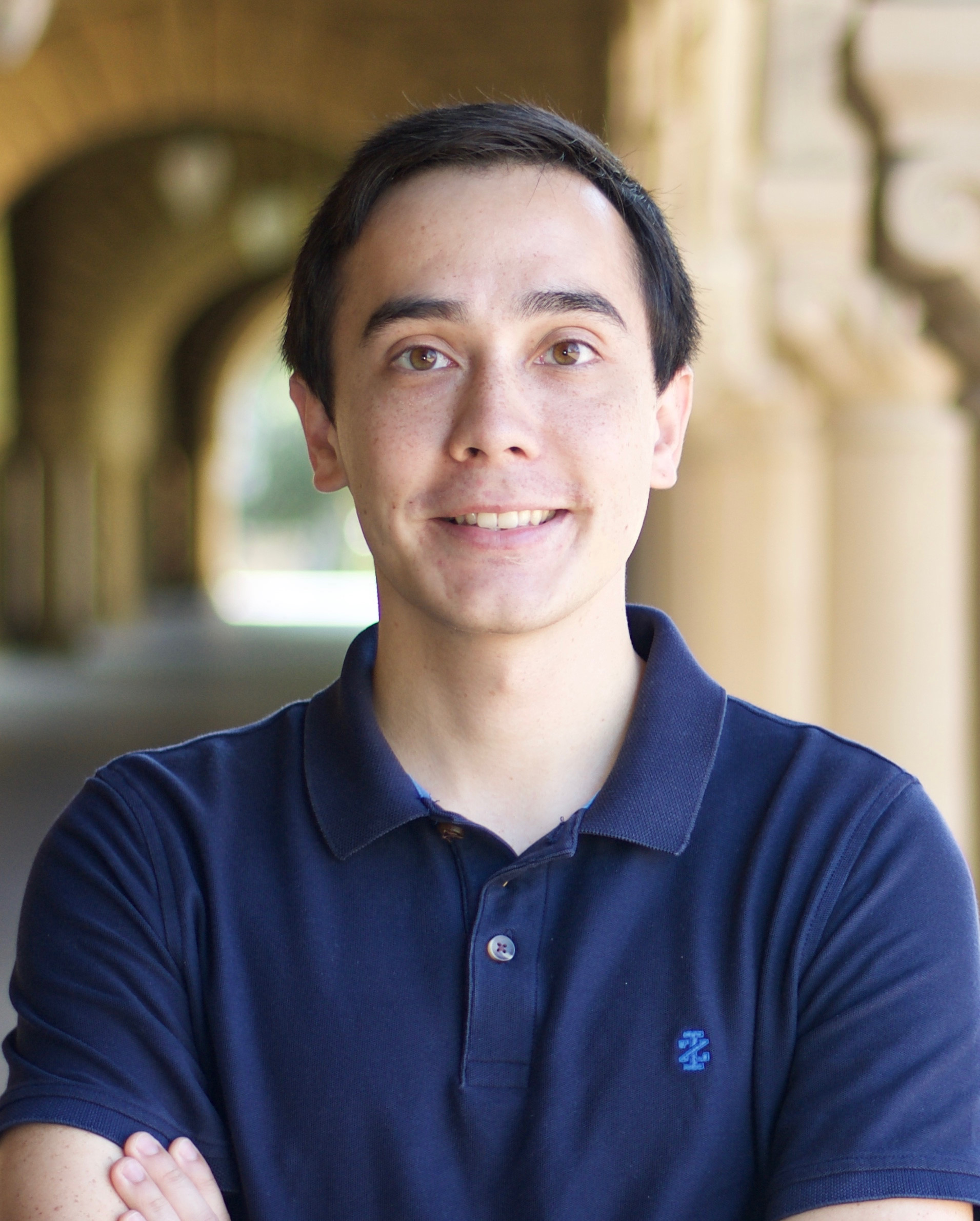}
is a Postdoctoral Scholar with the Space Rendezvous Laboratory at Stanford University. He received B.E. and B.S. degrees from the University of Western Australia, followed by M.S. and Ph.D. degrees in Aeronautics and Astronautics from Stanford University. His research focuses on autonomous angles-only navigation for distributed space systems, with recent contributions to the NASA Starling mission, achieving the first in-orbit demonstration of autonomous angles-only navigation for a satellite swarm. He received the M Charles Fogg Best Paper award at the 2021 IEEE Aerospace Conference, and was awarded Stanford University's William F. Ballhaus Prize and the Institute of Navigation's Bradley W. Parkinson prize in 2024 for his Ph.D. dissertation. 
\end{biographywithpic}

\begin{biographywithpic}{Ethan Foss}{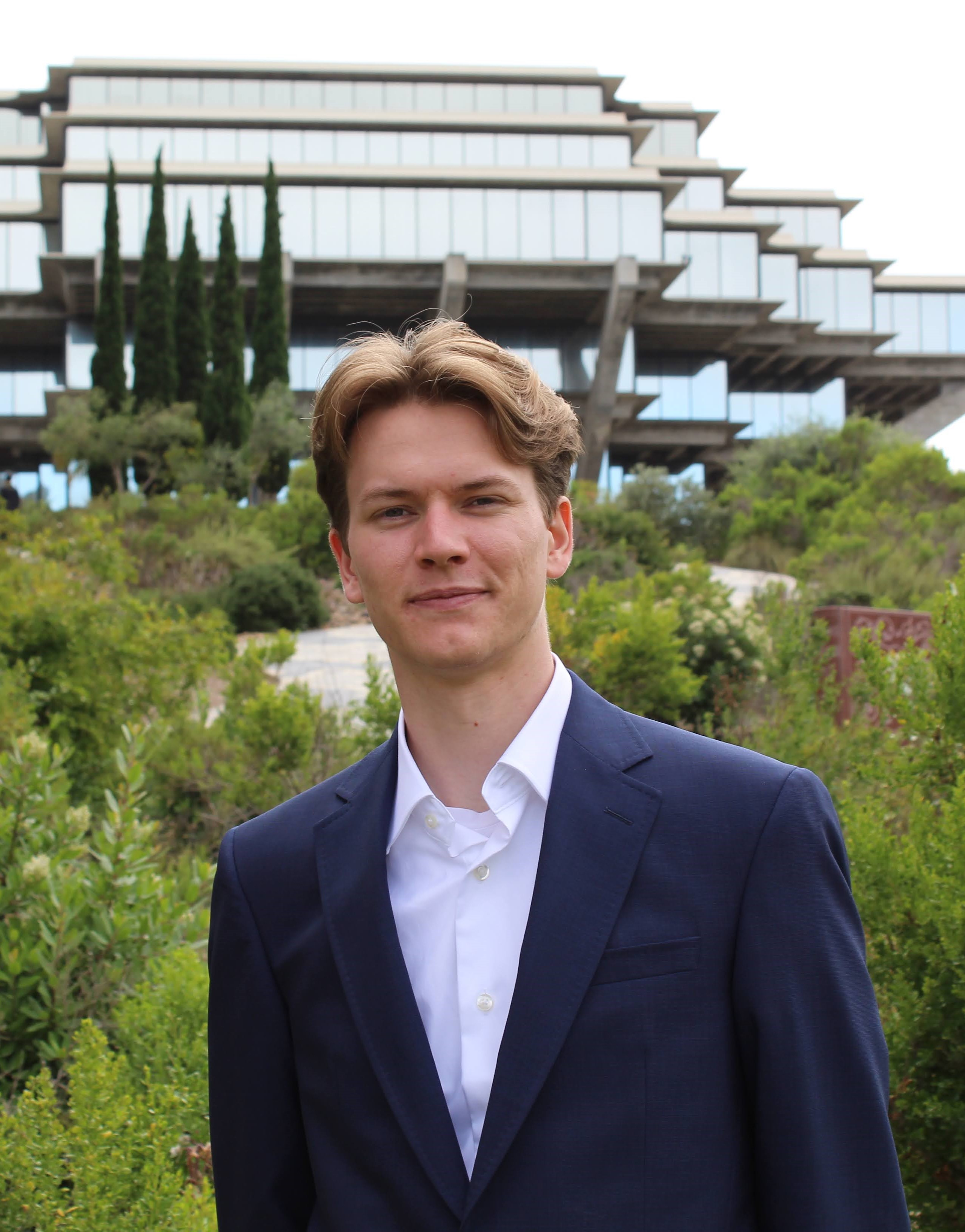}
Ethan Foss is a PhD student in the Space Rendezvous Laboratory at Stanford University. He graduated with a Bachelors degree and Masters degree in Mechanical Engineering from the University of California, San Diego in 2023 and 2024 respectively. His Master’s thesis work focused on the application of stochastic optimal control  for station keeping about quasi-periodic orbits in cislunar space. His research interests involve the integration of cutting-edge guidance and control algorithms with advanced spacecraft missions.
\end{biographywithpic}

\begin{biographywithpic}{Praneet Bhoj}{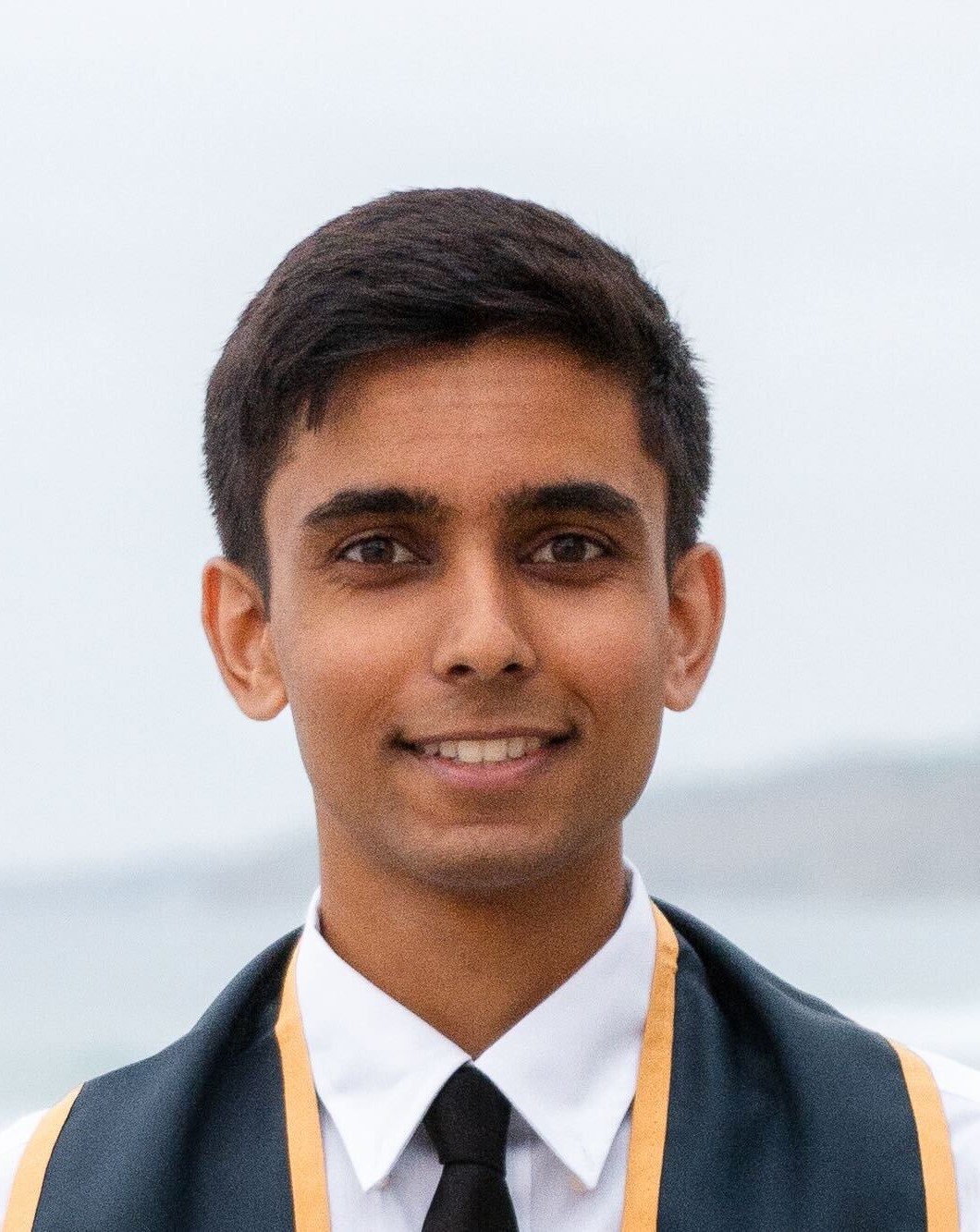}
Praneet Bhoj is a Master’s student in the Computer Science Department at Stanford University. He received his B.S. in Computer Science from UC San Diego in 2023. His research interests are in machine learning and foundation models for autonomous vehicles and spacecraft.
\end{biographywithpic}

\begin{biographywithpic}{Tommaso Guffanti}{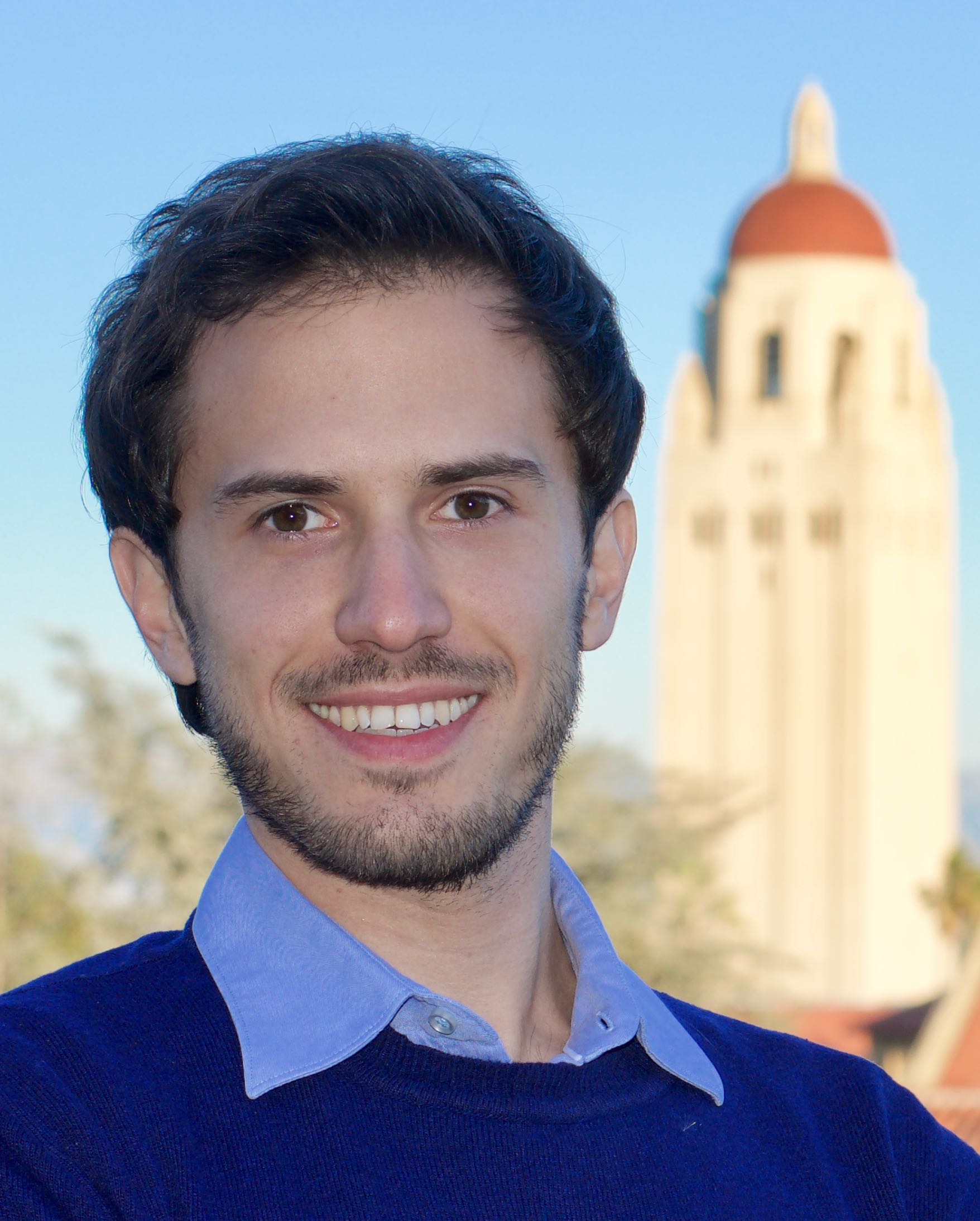}
Dr. Tommaso Guffanti is a Postdoctoral Scholar in the Space Rendezvous Lab at Stanford University. He received the B.S. and M.S. degrees in Aerospace engineering cum laude from Politecnico di Milano, and the Ph.D. degree in Aeronautics and Astronautics from Stanford University. Dr. Guffanti research contributions aim at developing cutting-edge guidance and control algorithms, and flight software to enable safe and autonomous functions and operations on-board space vehicles, in order to satisfy the requirements of the next generation of multi-spacecraft missions. During his doctorate and postdoctorate, he has been making research contributions in astrodynamics, safe and fault-tolerant multi-agent optimal control, and learning-based spacecraft motion planning and guidance for a variety of projects funded by national agencies and industry. He has over 15 scientific publications, including conference proceedings, and peer-reviewed journal articles. Dr. Guffanti is currently contributing to the GNC subsystem of the upcoming NSF-funded VISORS distributed telescopy mission, and leading the GNC team within the multi-institutional mission team. He has been awarded a doctoral Stanford Graduate Fellowship, a post-doctoral Stanford Center of Excellence for Aeronautics and Astronautics Scholarship, and a Stanford Emerging Technology Review Fellowship. He has been named excellent reviewer of the Journal of Guidance, Control, and Dynamics for three years.
\end{biographywithpic}

\begin{biographywithpic}{Simone D'Amico}{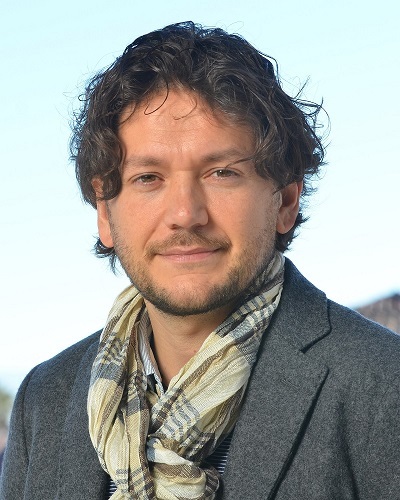}
Simone D’Amico is Associate Professor of Aeronautics and
Astronautics (AA), W.M. Keck Faculty Scholar in the School of Engineering, and Professor of Geophysics (by Courtesy). He is the Founding Director of the Stanford Space Rendezvous Laboratory, Co-Director of the Center for AEroSpace Autonomy Research (CAESAR), and Director of the Undergraduate Program in Aerospace Engineering at Stanford. He has 20+ years of experience in research and development of autonomous spacecraft and distributed space systems. He developed the distributed Guidance, Navigation, and Control
(GNC) system of several formation-flying and rendezvous
missions and is currently the institutional PI of four autonomous satellite swarms funded by NASA (STARLING, STARI) and by NSF (VISORS, SWARM-EX) with one of them
operational in orbit right now (Starling). Besides academia,
Dr. D’Amico is in the Advisory Board of four space start-ups
focusing on distributed space systems for future applications
in SAR remote sensing, orbital lifetime prolongation, and
space-based solar power. He was the recipient of several
awards, most recently the 2024 NASA Ames Honor Award
for the Starling mission, Best Paper Awards at IAF (2022),
IEEE (2021), AIAA (2021), AAS (2019) conferences, and the
M. Barry Carlton Award by IEEE (2020). He received the
B.S. and M.S. degrees from Politecnico di Milano (2003) and
the Ph.D. degree from Delft University of Technology (2010).
\end{biographywithpic}

\begin{biographywithpic}{Marco Pavone}{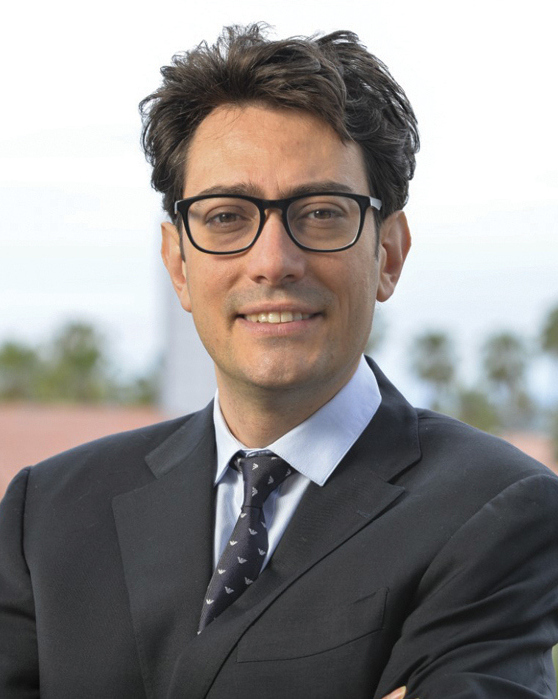}
Dr. Marco Pavone
is an Associate Professor of Aeronautics and Astronautics at Stanford University, where he is the Director of the
Autonomous Systems Laboratory and
Co Director of the Center for Automotive
Research at Stanford. Before joining
Stanford, he was a Research Technologist within the Robotics Section at the
NASA Jet Propulsion Laboratory. He
received a Ph.D. degree in Aeronautics and Astronautics from
the Massachusetts Institute of Technology in 2010. His main
research interests are in the development of methodologies
for the analysis, design, and control of autonomous systems,
with an emphasis on self-driving cars, autonomous aerospace
vehicles, and future mobility systems. He is a recipient of
a number of awards, including a Presidential Early Career
Award for Scientists and Engineers from President Barack
Obama, an Office of Naval Research Young Investigator
Award, a National Science Foundation Early Career (CAREER) Award, a NASA Early Career Faculty Award, and an
Early-Career Spotlight Award from the Robotics Science and
Systems Foundation. He was identified by the American Society for Engineering Education (ASEE) as one of America’s
20 most highly promising investigators under the age of 40.
His work has been recognized with best paper nominations
or awards at the European Control Conference, at the IEEE
International Conference on Intelligent Transportation Systems, at the Field and Service Robotics Conference, at the
Robotics: Science and Systems Conference, at the ROBOCOMM Conference, and at NASA symposia. He is currently
serving as an Associate Editor for the IEEE Control Systems
Magazine. He is serving or has served on the advisory
board of a number of autonomous driving start-ups (both
small and multi-billion dollar ones), he routinely consults
for major companies and financial institutions on the topic of
autonomous systems, and is a venture partner for investments
in AI-enabled robots
\end{biographywithpic}

\clearpage
\appendices{}
The appendices are home to the additional engineering details and explanations behind our approach to architecting {\fontfamily{cmtt}\selectfont Space-LLaVA} and our experimental results.

\section{Background: Space-LLaVA}
\subsection{Numerical Evaluation}\label{sec:appendix_pref_analysis}

\begin{figure}[b]
    \centering
    \includegraphics[width = \linewidth]{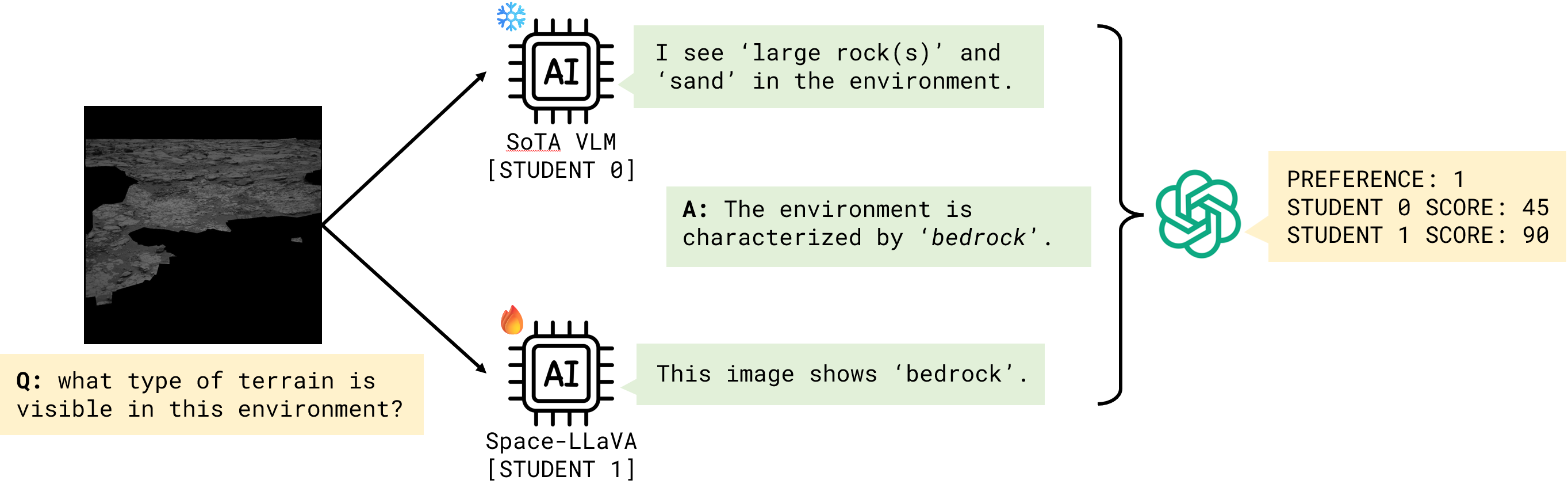}
    \caption{Preference analysis and numerical evaluation framework for GPT-assisted response characterization between {\fontfamily{cmtt}\selectfont Space-LLaVA} and a SoTA VLM, e.g., GPT-4o and base LLaVA. This diagram is meant to graphically depict our evaluation framework and does not represent real generations from {\fontfamily{cmtt}\selectfont Space-LLaVA} or GPT-4.}
    \label{fig:preference_analysis}
\end{figure}

We demonstrate the proficiency of {\fontfamily{cmtt}\selectfont Space-LLaVA} on select tasks through a comparison to existing pre-trained models. In this comparison, we leverage the GPT-4 language model as an automated text evaluator. Specifically, for every image-question pair in the evaluation dataset (withheld from training) for the particular task, we measure the relative performance of {\fontfamily{cmtt}\selectfont Space-LLaVA} against a zero-shot SoTA VLM. In this side-by-side comparison, each model's response is collected along with the expected, ground-truth answer. This tuple of three natural language responses is passed to GPT-4, prompted according to the following template, to determine which model's answer is most similar to the ground-truth response in terms of content and to score each candidate response on a scale from 0 (worst) to 100 (exemplar). In the following template, the base and {\fontfamily{cmtt}\selectfont Space-LLaVA} models are treated as the first and second student, respectively, and the prompt's overview is tailored to evaluate {\fontfamily{cmtt}\selectfont terrain comparison} as an emergent ability. If GPT-4 prefers the base model answer, it outputs 0; if it prefers the fine-tuned model answer, it outputs 1. 
We run this procedure over all withheld evaluation data for a particular task, and track each responses' numeric score and how many times the GPT-4 evaluator prefers the fine-tuned model answer. We provide the prompt's system message and user message below. We also visual depict this evaluation setup in \Cref{fig:preference_analysis}.

\begin{tcolorbox}[colback=gray!40, colframe=gray!40, sharp corners]
{\fontfamily{cmtt}\selectfont 
System Message
\newline
\newline
You will act as the judge of natural language responses from two students in planetary science. You will be presented with a 'QUESTION', the desired 'GROUND-TRUTH' answer, and the responses from two students to be evaluated. Your job is to score each response and decide which of the two answers is most similar to the 'GROUND-TRUTH' response based on the response's content, i.e., disregard whether a response simply has a similar structure to the 'GROUND-TRUTH' answer.}

\end{tcolorbox}

\begin{tcolorbox}[colback=gray!40, colframe=gray!40, sharp corners]
{\fontfamily{cmtt}\selectfont 
User Message
\newline
\newline
OVERVIEW: 
Two graduate students in planetary science are presented with an open-ended `QUESTION' which evaluates each student's ability to compare and contrast the characteristics of at least two different terrain types in a camera image of Mars' landscape. Your job is to score each student's response with a numeric grade and determine which student's response is most similar to the `GROUND-TRUTH' response.
\newline
\newline
RULES:

1) In your response, you will return three scores, i.e., a PREFERENCE score, a numeric score for the response from `STUDENT 0', and a numeric score for the response from `STUDENT 1'.

2) The PREFERENCE score should be as a single number corresponding to the student with the answer most similar in content and meaning to the `GROUND-TRUTH' answer, e.g., you should return a 0 if `STUDENT 0' is preferable to `STUDENT 1', and conversely, you should return a 1 if `STUDENT 1' is more preferable to `STUDENT 0'.

3) The score you give to `STUDENT 0' and `STUDENT 1' should be an integer number between 0 (worst) and 100 (exemplar) reflecting the degree of similarity between the student's response with the `GROUND-TRUTH' answer. For example, a very similar response to the ground-truth answer should receive a high score.

4) If you are unsure which student's response is preferable or the exact numeric grade to assign either student, please use your best judgment.

5) Give your answer in the following format:

PREFERENCE: <YOUR PREFERENCE SCORE HERE>

STUDENT 0 SCORE: <YOUR SCORE FOR STUDENT 0 HERE>

STUDENT 1 SCORE: <YOUR SCORE FOR STUDENT 1 HERE>

6) Finally, strictly follow the format above and do not provide an explanation to justify your evaluation.
\newline
\newline
CONTENT:

Question: [QUESTION]

`GROUND-TRUTH' answer: [GROUND-TRUTH]

`STUDENT 0' response: [STUDENT 0]

`STUDENT 1' response: [STUDENT 1]}

\end{tcolorbox}

\subsection{Space-LLaVA Dataset}

\subsubsection{The AI4Mars and MICD Dataset}\label{sec:appendix_data_augmentation}
As discussed in \Cref{sec:ai4mars_scoti}, for each sample in the MSL AI4Mars dataset, we synthetically generate a language annotation in a, e.g., instruction-following, conversation format to accomplish one of the following seven objectives, as shown in \Cref{fig:spacellava-composition}:
\begin{enumerate}[wide, labelwidth=1pt, labelindent=4pt]
    \item {\fontfamily{cmtt}\selectfont scene description}
    \item {\fontfamily{cmtt}\selectfont terrain description}
    \item {\fontfamily{cmtt}\selectfont terrain classification}
    \item {\fontfamily{cmtt}\selectfont terrain comparison}
    \item {\fontfamily{cmtt}\selectfont grain characterization}
    \item {\fontfamily{cmtt}\selectfont terrain localization}
    \item {\fontfamily{cmtt}\selectfont group terrain localization.}
\end{enumerate}
We translate the segmentation masks from the AI4Mars dataset into a color-codes superimposed on the raw image, \Cref{fig:ai4mars_dataset_curation}, from Curiosity to provide a SoTA VLM, i.e., GPT-4o, with \emph{visual context} identifying and localizing terrain types in the image. Hence, we query GPT-4o with the augmented AI4Mars image and additional context for annotation provided in natural language through the user prompt, e.g., we match each color to the four terrain types classified by AI4Mars, among other context clues. The exact context that we provide in the prompt varies slightly between each fine-tuning task; accordingly, for brevity, we provide our user and system message template below for one fine-tuning task. Given we wish to generate an annotation for {\fontfamily{cmtt}\selectfont terrain description}, as in \Cref{fig:ai4mars-terrain-desc}, we provide the following user and system message, replacing the {\fontfamily{cmtt}\selectfont [QUESTION]} with a randomly selected question from a choice of ten, as described in \Cref{sec:ai4mars_scoti}, to GPT-4o along with a color-coded AI4Mars image. 

\begin{tcolorbox}[colback=gray!40, colframe=gray!40, sharp corners]
{\fontfamily{cmtt}\selectfont 
User Message
\newline
\newline
I would like to build a dataset of Visual-Question-Answer (VQA) tuples based on the AI4Mars dataset - a dataset for terrain classification on Mars' surface - in order to train a vision-language model to perform the task of `terrain description'.
\newline\newline
You will be presented with a color-coded image from the AI4Mars dataset and your task will be to describe one of the terrain types visible to you. Specifically, the image will be derived from a grayscale image taken onboard a Martian rover highlighting one or more terrain types of interest on Mars, e.g., `regolith', `bedrock', `sand' or `large rock(s)', with color codes to inform your analysis.
\newline\newline
For your reference, a `Blue' highlight corresponds to the `regolith' terrain type; a `Green' highlight corresponds to the `bedrock' terrain type; a `Red' highlight corresponds to the `sand' terrain type; a `Purple' highlight corresponds to the `large rock(s)' terrain type. The color codes are meant to inform you which terrain types are visible and where, generally speaking, each terrain is located in the image. In your response, describe the requested terrain type without mentioning its color. For example, the color `Red' indicates to you that `sand' is both 1) present in the image and 2) located in a particular region of the image, which you should use in your response but do not mention that the `sand' is colored `Red' or that it is highlighted. In your response, please only analyze the terrain types highlighted through color codes and focus your analysis on the terrain type requested in the following question.
\newline\newline
Given this image with color-coded terrain, your task is to [QUESTION]}
\end{tcolorbox}

\begin{tcolorbox}[colback=gray!40, colframe=gray!40, sharp corners]
{\fontfamily{cmtt}\selectfont 
System Message
\newline
\newline
You are a helpful assistant and provide short, concise responses with specific terrain features.}
\end{tcolorbox}

Conversely, for the MICD, shown in \Cref{fig:micd-sample}, we required a diverse set of prompts which elicit an image caption from a camera capture of Mars' landscape. Therefore, we randomly assign each MICD image one of the following ten image caption prompts:

\begin{enumerate}[wide, labelwidth=1pt, labelindent=4pt]
    \item \small\selectfont``Provide a short caption for this image."
    \item ``Summarize the relevant features in your view."
    \item ``Describe the image shown."
    \item ``What is in this image?"
    \item ``Given the image shown, write a caption."
    \item ``Give a brief description of the contents in this image."
    \item ``Write a concise caption to reflect the contents of the image."
    \item ``What is present in this view?"
    \item ``What is a short description for the visible scene?"
    \item ``Summarize this photo with a caption.".
\end{enumerate}

\subsubsection{SpaceScienceQA Dataset}\label{sec:appendix_scienceQA}

As discussed in \Cref{sec:space-science-QA}, we use a two-step QA generation pipeline to first architect QA pairs from publications in astrophysics with assistance from a SoTA LLM, i.e., GPT-4, and then cleanse the dataset of low-quality and low-information pairs before fine-tuning. Specifically, in QA generation, we encourage GPT-4 to adopt the persona of a graduate school professor designing an exam based on a sliding window of passages from an assigned research paper. We then outline the spirit and rules for programmatic QA generation through the user prompt provided below.

\begin{tcolorbox}[colback=gray!40, colframe=gray!40, sharp corners]
{\fontfamily{cmtt}\selectfont 
User Message\newline\newline
OVERVIEW: You will be provided a short, 1800 character PASSAGE from a recent research paper in the category of `[CATEGORY]', and based on the PASSAGE, your task is to curate a sophisticated, self-contained question designed to evaluate one's understanding of the scientific content in the PASSAGE. Once you have formed the question, you will then provide the question's answer, to the best of your abilities, using the information presented in the PASSAGE, too. The paper is titled `[TITLE]' with the following abstract: `[ABSTRACT]'\newline\newline
RULES:\newline
1) Please ensure that the question you generate includes all the necessary context in order to answer it, i.e., one should be able to fully understand the question without access to the research paper. \newline
2) Your answer should be clear, specific and leverage comprehensive information based on the provided PASSAGE. This answer should require concepts from the research paper. \newline 
3) Do NOT mention the PASSAGE in the question or answer. Also, each question and answer pair should be understandable to someone without direct access to the PASSAGE. For example, each question and answer pair should not include any external references.\newline
4) Limit the answer you provide to three to five sentences.\newline
5) Provide your question and answer in the following response format:\newline
[QUESTION]: <YOUR QUESTION HERE>.\newline
[ANSWER]: <YOUR ANSWER HERE>.\newline\newline
PASSAGE: `[PASSAGE]'.}
\end{tcolorbox}

\begin{tcolorbox}[colback=gray!40, colframe=gray!40, sharp corners]
{\fontfamily{cmtt}\selectfont 
System Message\newline\newline
You will assume the role of a graduate school professor in `[CATEGORY]' with a deep and well-versed understanding of your field. You are preparing the questions and answers for an exam in your graduate-level class based on a research paper assigned as mandatory reading. Accordingly, the exam you generate should be sophisticated enough for advanced, graduate-level students. Importantly, while you will select question and answer pairs based on passages from the paper, your question and answer pairs should never reference specific figures or equations in the paper. Instead, your exam of question and answer pairs should focuse on the conceptual aspects of the passage.}
\end{tcolorbox}

As discussed in \Cref{sec:space-science-QA}, we populate the prompt templates above with 1,000 publications evenly distributed across five specializations in arXiv's astrophysics category: Cosmology and Nongalatic Astrophysics; Earth and Planetary Astrophysics; Astrophysics of Galaxies; Instrumentation and Methods for Astrophysics; and Solar and Stellar Astrophysics. For each candidate QA pair, we query GPT-4 again to ascertain the consistency of the soon-to-be ground-truth answer with respect to the original passage on a numeric scale from 0 (worst) to 100 (exemplar)---providing a measure of \emph{answer quality}. Separately, we query Vicuna-13B-v1.5, the language backbone of LLaVA-v1.5-13B, to answer the candidate question zero-shot, i.e., without access to the original passage, which is also scored by GPT-4 on the numeric scale described previously---providing a measure of the answer's \emph{information gain} through fine-tuning. Importantly, we reject all QA pairs from the dataset whose ground-truth answer scores below a 90 out of 100 points and QA pairs for which Vicuna scores within at least 20 points of the ground-truth answer, ensuring both high-quality and informative QA pairs. We independently score each response, i.e., the ground-truth and zero-shot answer, with the following user and system message passed to GPT-4.

\begin{tcolorbox}[colback=gray!40, colframe=gray!40, sharp corners]
{\fontfamily{cmtt}\selectfont 
User Message\newline\newline
OVERVIEW: You will be provided a short, 1800 character PASSAGE from a recent research paper in the category of `[CATEGORY]' and your task is to grade an answer from one of your students with a numeric score from 0 (worst) to 100 (exemplar) based on accuracy and fidelity with respect to the PASSAGE. The paper is titled `[TITLE]' with the following abstract: `[ABSTRACT]'\newline\newline
RULES:\newline
1) Your numeric score should be an integer between 0 (worst) and 100 (exemplar), i.e., do not include a decimal in your grade.\newline
2) A high numeric score should reflect a response demonstrating high accuracy with respect to the conceptual and scientific information provided in the PASSAGE.\newline
3) A low numeric score should be assigned to a response which is not reflective of or does not use the scientific information and concepts presented in the PASSAGE.\newline
4) You should also assign a low score to an answer if the answer OR the question makes a reference to the existence of the PASSAGE itself, i.e., the question and answer should be understandable to someone unable to read the PASSAGE.\newline
5) If you are unsure, then please use your best judgment.\newline
6) Provide your numeric score in the following response format:\newline
[GRADE]: <YOUR NUMERIC SCORE HERE>\newline
7) Do not provide an explanation for your grade. Only return the numeric score as an integer.\newline\newline
PASSAGE: `[PASSAGE]'\newline
QUESTION: `[QUESTION]'\newline
STUDENT RESPONSE: `[ANSWER]'.}
\end{tcolorbox}

\begin{tcolorbox}[colback=gray!40, colframe=gray!40, sharp corners]
{\fontfamily{cmtt}\selectfont 
System Message\newline\newline
You will assume the role of a graduate school professor in `[CATEGORY]' with a deep and well-versed understanding of your field. You have recently administered an open response exam to your graduate-level class and now you are grading each individual response on a numeric scale from 0 (worst) to 100 (exemplar).}
\end{tcolorbox}

\section{Background: On-Orbit Imagery Evaluation}\label{sec:appendix_orbital_space}

\subsection{Data Sources}

As discussed in Section \ref{sec:orbit}, we collate a test set of satellite flight images in order to evaluate FM performance on a dissimilar but space-oriented task. A summary of image characteristics is presented in Table \ref{table:satellitedataset}.

\begin{table*}[ht]
\centering
\begin{tabular}{lllll}
\toprule
Camera Source     & \# Images & Resolution                      & Color     & Targets      \\ \midrule
Mango Cam. 0      & 566       & 2048 $\times$ 2048 & RGB       & Resolved     \\
Mango Cam. 3      & 313       & 752 $\times$ 580   & Grayscale & Resolved     \\
Starling 2 Cam. 1 & 282       & 1280 $\times$ 1024 & Grayscale & Non-resolved \\
Starling 2 Cam. 2 & 610       & 1280 $\times$ 1024 & Grayscale & Non-resolved \\
ExoRomper & 1234      & 1296 $\times$ 972  & RGB       & Resolved    \\
\bottomrule
\end{tabular}
\caption{Image sources and properties in the satellite imagery test dataset.}
\label{table:satellitedataset}
\end{table*}

The first image source is the \textit{Mango} spacecraft, launched in 2010 as part of the PRISMA mission \cite{damico_argon_2013}. A single observer satellite (`Mango') obtained images of a noncooperative target satellite (`Tango') in low Earth orbit. During the extended PRISMA mission, Mango performed a rendezvous with Tango down to an inter-satellite distance (ISD) of two meters, meaning the target object is resolved. An example image, along with the prompt provided to the FM, is given in Figure \ref{fig:prisma_example}.

\begin{figure}[htb]
     \centering
     \includegraphics[width = \textwidth]{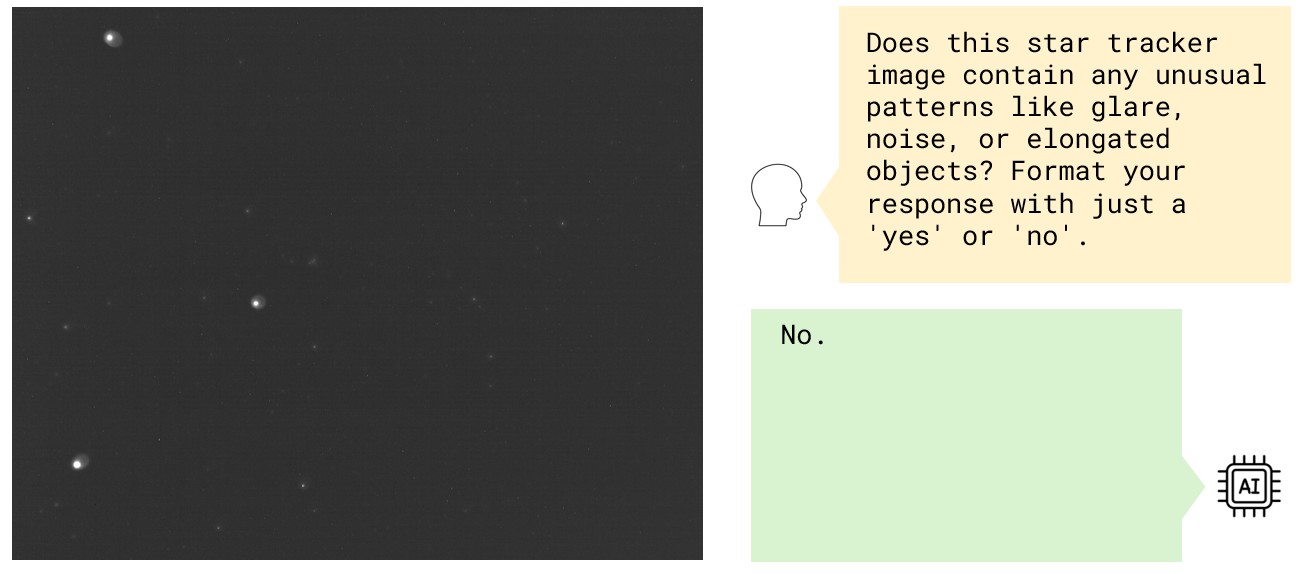}
     \caption{An example \textit{Starling} image and corresponding FM prompt.}
     \label{fig:starling_example}
\end{figure}

The second image source is the \textit{Starling} SV2 CubeSat, launched in 2023 as part of the NASA Starling swarm \cite{kruger_flight_2024}. Each of the four Starling CubeSats carries two star trackers used to perform simultaneous attitude determination and angles-only orbit determination by measuring the bearing angles to other swarm members. Typical ISDs during the mission were tens to hundreds of kilometers, meaning target objects are unresolved. An example image, along with the prompt provided to the FM, is given in Figure \ref{fig:starling_example}.

\begin{figure}[htb]
     \centering
     \includegraphics[width = \textwidth]{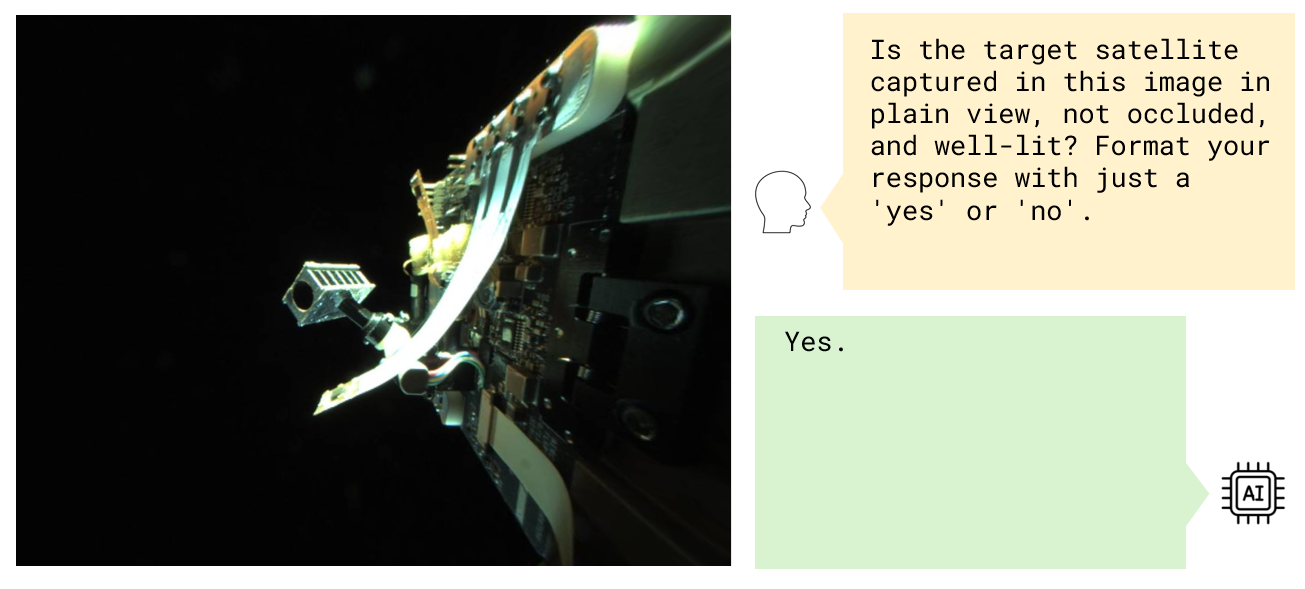}
     \caption{An example \textit{ExoRomper} image and corresponding FM prompt.}
     \label{fig:exoromper_example}
\end{figure}

The third image source is the \textit{ExoRomper} dataset, produced by The Aerospace Corporation. The dataset was assembled in support of Aerospace's Slingshot program \cite{weiher_exoromper_2022}, which launched the 12U Slingshot 1 CubeSat in 2022. The ExoRomper payload produces imagery of a maneuverable 3U CubeSat model (along with pose information) using thermal and visible light cameras. An example image, along with the prompt provided to the FM, is given in Figure \ref{fig:exoromper_example}.

\subsection{Data Curation}

\begin{figure*}[htb!]
    \centering
    \includegraphics[width = \textwidth]{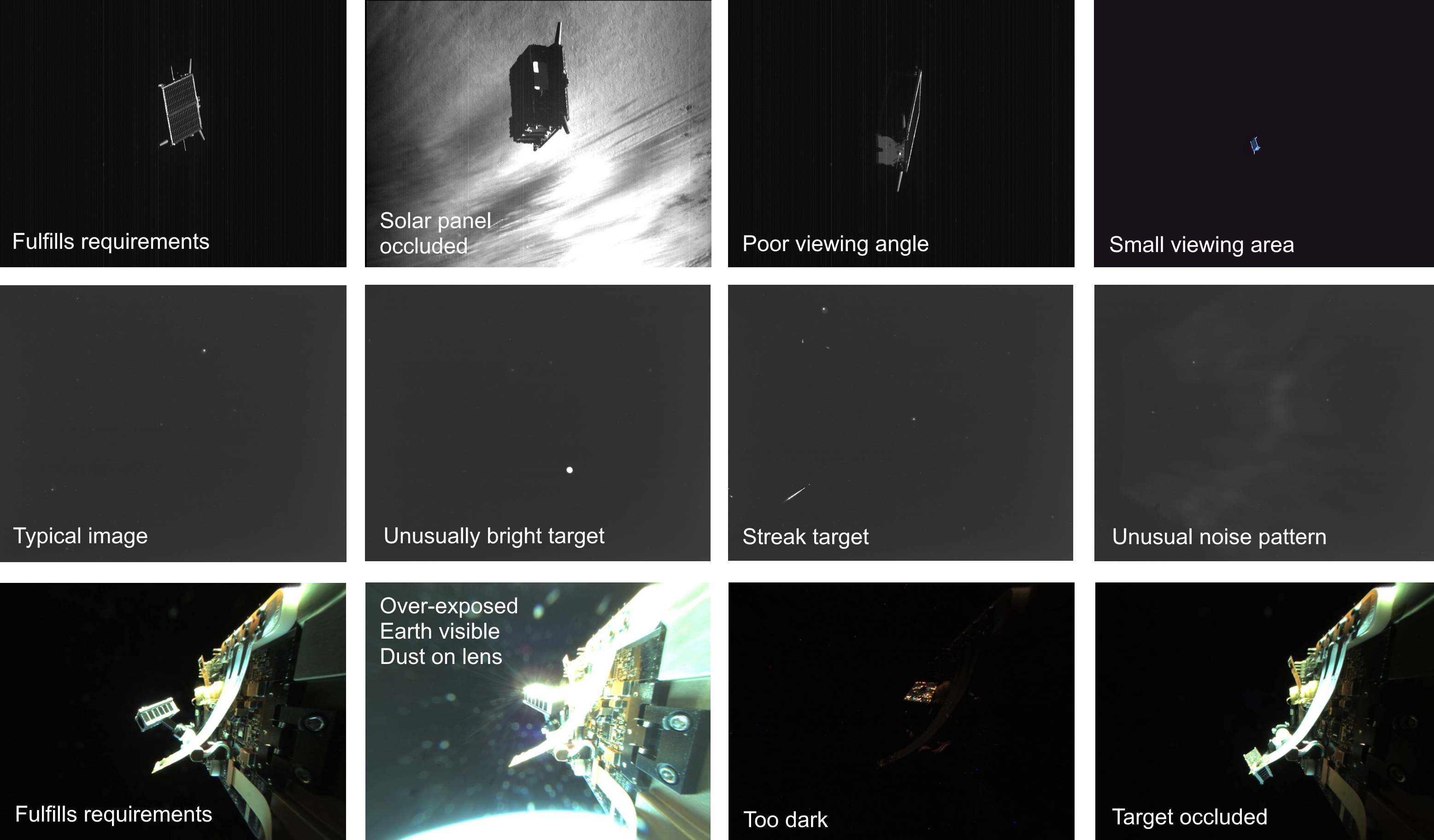}
    \caption{Examples of images which fulfill or do not fulfill curation requirements for \textit{PRISMA} (top row), \textit{Starling} (middle row) and \textit{ExoRomper} (bottom row).}
    \label{fig:satellite_examples}
\end{figure*}

We hand-label images with whether they do or do not fulfill the stated user requirements, as listed below. Canonical examples of images which fulfill or do not fulfill the requirements are presented in Figure \ref{fig:satellite_examples}. Of the 879 PRISMA images, 65 images were hand-labeled as fulfilling the requirements; of the 892 Starling images, 380 images fulfilled the requirements; and of the 1234 ExoRomper images, 361 images fulfilled the requirements.

It is important to qualify issues with the experimental setup that negatively impact model performance for this task. Though the datasets were classified manually in a binary manner, in reality the classes display a high degree of potential ambiguity and overlap. It is therefore unreasonable to expect any VLM to achieve perfect classification performance, which necessitates the need for fine-tuning or few-shot prompting when such ambiguity is present. However, since the primary aim of this evaluation is to demonstrate that {\fontfamily{cmtt}\selectfont Space-LLaVA} does not lose generality after fine-tuning procedure, further experimentation in this regard is beyond the scope of this paper and can be explored in future iterations. 


\begin{tcolorbox}[colback=gray!40, colframe=gray!40, sharp corners]
{\fontfamily{cmtt}\selectfont 
Curation Requirements for the Satellite Dataset\newline\newline
1) Select PRISMA images which provide a clear view of the target's solar panel, such that they could be used to assess potential exterior damage. Consider the following:\newline
- Occlusion of the solar panel.\newline
- Angle with respect to the camera.\newline
- Adequate visual size.\newline
- Adequate lighting.\newline
- Presence of corrupted image data.\newline
2) Select Starling images which possess unusual properties atypical of a star tracker image. Examples of unusual properties include: \newline
- One object much brighter than all others.\newline
- Streaks or other elongated objects.\newline
- Strong sun glare or glints.\newline
- Irregular noise patterns.\newline
3) Select ExoRomper images that provide a clear view of the target. Consider the following:\newline
- The target should not occlude or be occluded by other apparatus, i.e. is silhouetted against a black background.\newline
- Earth is not in the background.\newline
- Target is well-lit and not over-exposed.\newline
- Target is in focus.\newline
- No excessive dust or noise on the lens.
}
\end{tcolorbox}

\section{Background: Lunar Simulation}
\label{sec:appendix_lunar_sim}
In what follows, we present additional results for all experimental scenarios conducted within the lunar simulation environment. Specifically, we provide: (1) the base user message---which, apart from the ground team's path coordinates, is unchanged across scenarios---and (2) FM-generated outputs along with visualizations of the scenarios.

\textbf{User message:}
\begin{tcolorbox}[colback=gray!40, colframe=gray!40, sharp corners]
{\fontfamily{cmtt}\selectfont 
User Message
\newline
\newline
You are the onboard intelligence of a lunar rover operating on the lunar surface. You are provided with four images, in this order: three onboard cameras from the rover (Front, Front Left, Front Right) and one Top-Down View of the lunar surface where the rover is located. The top-down view of the lunar surface is equipped with a grid overlay to assist with navigation. The grid consists of blue lines representing cell divisions. Each grid cell is 1m x 1m. The rover’s current position is `[R-POSITION]' and its location is highlighted as a yellow rectangle in the top-down view. The lander is located at lunar coordinates `[L-POSITION]' and its location is highlighted as a large red rectangle in the top-down view.\newline
\newline
Your primary tasks are:
\newline
1.	Scene Analysis: Monitor the rover’s state and accurately describe the scene using inputs from multiple cameras (Front, Front Left, Front Right, Left, Top-Down View). When analyzing the top-down view, focus on making connections with what you observe from the onboard cameras.
\newline
2.	Plan Evaluation: Evaluate navigation plans sent by the ground team to ensure they are safe, feasible, and efficient.
\newline
3.	Suggestions: If the ground plan has risks or inefficiencies, propose alternative navigation plans with justifications.\newline
\newline
**Scene Analysis:**
\newline
You are provided with three images from the rover’s cameras (Front, Front Left, Front Right). Focus on describing the scene using only data from the rover’s cameras. Make sure to refer to possible hazards and shadows visible from the cameras, specifying in which parts of the image these hazards might be visible (e.g., right, left, center, etc.): make sure this visual localization is consistent with the successive plan evaluation and potential definition of an alternative plan. Address the following: CONTINUED...
}
\end{tcolorbox}

\begin{tcolorbox}[colback=gray!40, colframe=gray!40, sharp corners]
{\fontfamily{cmtt}\selectfont 
...CONTINUED
\newline
- Terrain Features: Identify rocks, craters, slopes, or flat regions.
\newline
- Lighting Conditions: Pay particular attention to areas with shadows. Consider how shadows might obscure the terrain and present visibility challenges.
\newline
- Navigation-Relevant Features: Note clear paths, hazards, or obstacles critical to navigation.
\newline\newline
**Plan Evaluation:**
The ground team has proposed a plan for the rover to navigate toward the lander, heading towards the right of the rover. The proposed waypoints are:\newline
[GT WAYPOINTS]\newline
The first 3 waypoints are visualized as orange rectangles in the top-down view, in accordance with their coordinate values.
\newline\newline
When evaluating the plan, consider the following criteria:\newline
1.	Safety: Identify any hazards in the proposed path (e.g., obstacles, craters, or terrain features that might damage or immobilize the rover).\newline
2.	Feasibility: Assess whether the rover can traverse the path based on the terrain features and constraints visible in the images.\newline
3.	Efficiency: Favor paths with lower energy consumption, assuming all safety requirements are met.
\newline\newline
If the proposed plan presents significant risks or inefficiencies, suggest an alternative navigation plan that avoids these hazards. Your suggestions must include:\newline
CONTINUED...
}
\end{tcolorbox}

\begin{tcolorbox}[colback=gray!40, colframe=gray!40, sharp corners]
{\fontfamily{cmtt}\selectfont 
...CONTINUED
\newline
- Alternative Waypoints: Provide a set of waypoint coordinates explicitly defined using the grid system from the top-down view. This ensures the waypoints are consistent with the spatial layout.\newline
- Path Description: Offer a clear textual explanation of the proposed path.\newline
- Justification: Provide a concise rationale for the suggested changes, addressing safety, feasibility, and efficiency considerations.\newline
\newline
Response Structure:
\newline
1.	Scene Analysis: Describe the terrain and conditions based on the images.\newline
2.	Plan Evaluation: Analyze the ground team’s proposed plan and highlight concerns or benefits, particularly regarding shadowed areas.\newline
3.	Alternative Suggestions: Propose an alternative path if needed, with justifications focusing on shadowed areas, safety, feasibility, and efficiency.}
\end{tcolorbox}

\subsection{Hazardous terrain and uneven regolith}
\textbf{Outputs and visualizations:}

\begin{tcolorbox}[colback=gray!40, colframe=gray!40, sharp corners]
{\fontfamily{cmtt}\selectfont 
FM Output
\newline
\newline
SCENE ANALYSIS
\newline
Based on the images provided by the rover's cameras (Front, Front Left, Front Right) and the top-down view, here's a detailed analysis of the lunar surface surrounding the rover:\newline
\newline
**Front Camera:**\newline
- Terrain Features: A large rock is noticeable on the left side of the image. The terrain appears uneven with some slopes.\newline
- Lighting Conditions: The image shows good illumination with parts of the terrain casting shadows, especially near the large rock.\newline
- Navigation-Relevant Features: The large rock on the left could be an obstacle. The middle section seems relatively clear, but the uneven terrain and shadows suggest caution.\newline
\newline
**Front Left Camera:**\newline
- Terrain Features: The large rock remains visible on the far-left side, and the ground has small undulations.\newline
- Lighting Conditions: Similar to the front camera, shadows are cast by the rock and the terrain undulations. Overall, the lighting is decent.\newline
- Navigation-Relevant Features: The path straight and slightly to the left appears to have obstacles like rocks and uneven ground. The rock on the far-left corner is a notable obstacle.\newline
\newline
**Front Right Camera:**\newline
- Terrain Features: The large rock is less prominent here, creating a clearer path forward. Small rocks and uneven patches are visible.\newline
- Lighting Conditions: Good lighting, but shadows exist near small rocks and terrain features.\newline
- Navigation-Relevant Features: Compared to the front and front-left views, a slightly clearer path exists forward and to the right, but caution is still needed due to uneven patches.\newline
\newline
**Top-Down View:**\newline
- Current Position: The rover is located at (3, 19), highlighted in yellow.\newline
- Lander Position: The lander is at (0,0), highlighted in red.\newline
- Proposed Waypoints: The waypoints proposed by the ground team are in orange rectangles at (0.5, 16), (0.5, 12), and (0, 8).\newline
- Visibility of Hazards: The top-down view highlights dark regions and varying terrain features that may pose navigation risks.\newline
\newline
CONTINUED...
}
\end{tcolorbox}

\begin{tcolorbox}[colback=gray!40, colframe=gray!40, sharp corners]
{\fontfamily{cmtt}\selectfont 
...CONTINUED
\newline
PLAN EVALUATION
\newline
Ground Team's Proposed Plan:\newline
- Proposed Path: The rover is to navigate from (3, 19) to (0.5, 16), (0.5, 12), (0., 8), and finally to (0, 0).\newline
- Safety:\newline
  1. From (3, 19) to (0.5, 16): The route passes through darker shaded areas suggesting potential shadows and hidden obstacles.\newline
  2. From (0.5, 16) to (0.5, 12): Traverses through further shadowed regions increasing risks of unseen obstacles.\newline
  3. From (0.5, 12) to (0, 8): The path involves navigating close to uneven and potentially hazardous terrain.\newline
- Feasibility: Given the terrain observed from the images, the rover may face challenges navigating through shadowed and uneven regions.\newline
- Efficiency: The path is direct but needs reassessment due to potential hazards. An efficient path should balance safety and minimal energy consumption while avoiding hazards.\newline
\newline
Alternative Suggestions\newline

Alternative Path:\newline
Given the concerns highlighted above, here is an alternative path with defined waypoints:\newline

1. From (3, 19) to (5, 18)\newline
2. From (5, 18) to (6, 15)\newline
3. From (6, 15) to (4, 11)\newline
4. From (4, 11) to (2, 8)\newline
5. From (2, 8) to (0, 0)\newline
\newline
Path Description:\newline
- Step 1: (3, 19) to (5, 18): This detour provides a safer path through an illuminated region avoiding the large rock and steep slopes observed.\newline
- Step 2: (5, 18) to (6, 15): Continue maneuvering through clearer terrain. This avoids the central shadowed area.\newline
- Step 3: (6, 15) to (4, 11): The path balances between shadowed regions and more illuminated areas, aiming to minimize risk.\newline
- Step 4: (4, 11) to (2, 8): This segment navigates through relatively clearer regions based on the top-down grid view.\newline
- Step 5: (2, 8) to (0, 0): The final stretch leading to the lander while avoiding the most hazardous terrain.\newline

Justification:\newline
- Safety: This path avoids darker, shadowed regions and potential hazards observed in the images and top-down view. It maintains a continuous path through more illuminated regions, ensuring better visibility.\newline
- Feasibility: Given the terrain and lighting conditions, these waypoints provide a balance of clear terrain and feasible navigation.\newline
- Efficiency: Despite being longer, the chosen path minimizes potential energy consumption risks due to immovable obstacles and provides a balanced approach to navigate safely and efficiently.\newline

This alternative route ensures a safer approach by minimizing shadowed regions and potential unseen hazards while offering a balanced way to navigate towards the lander.
}
\end{tcolorbox}

\subsection{Large obstacles and structural threats}

\textbf{Outputs and visualizations:}
\begin{figure}[ht!]
    \centering
    \begin{subfigure}{\linewidth}
        \centering
        \includegraphics[width=\linewidth]{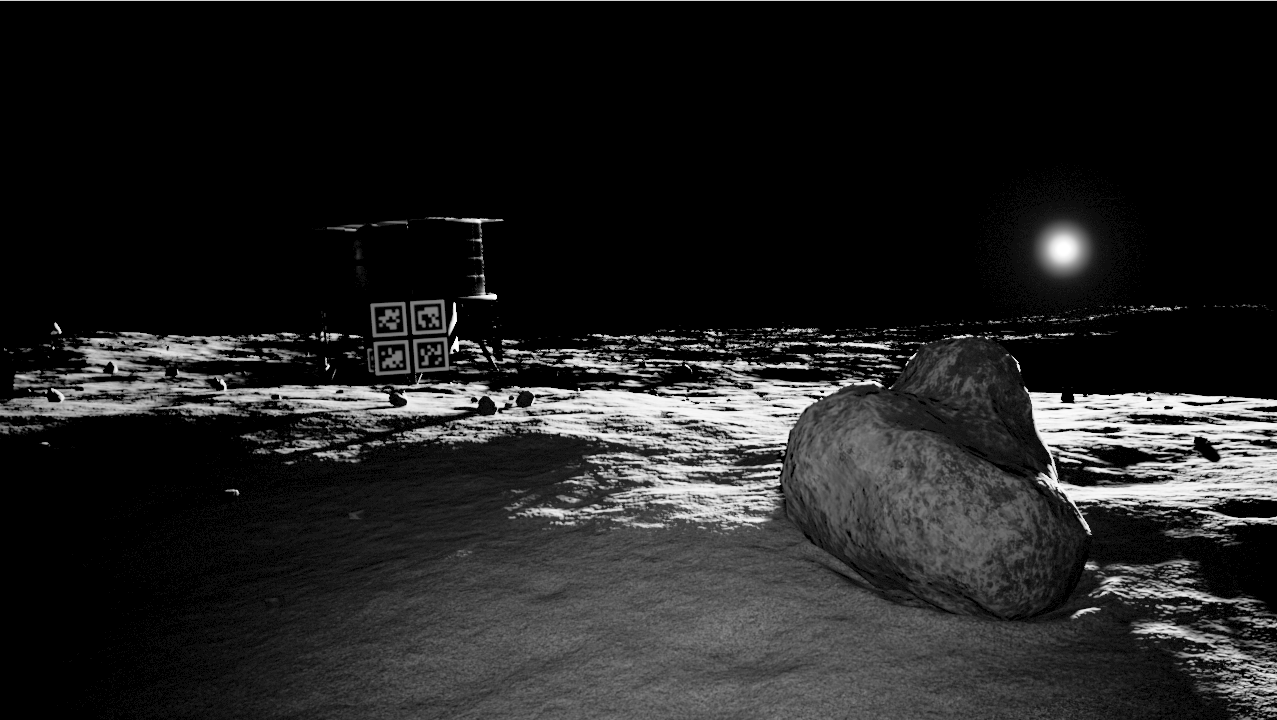}
        \caption{Onboard image provided to GPT-4.}
        \label{fig-lun-sim:front-scenario-2}
        \vspace{1.25em}
    \end{subfigure}
    \vspace{0.25em}
    \begin{subfigure}{\linewidth}
        \centering
        \includegraphics[width=\linewidth]{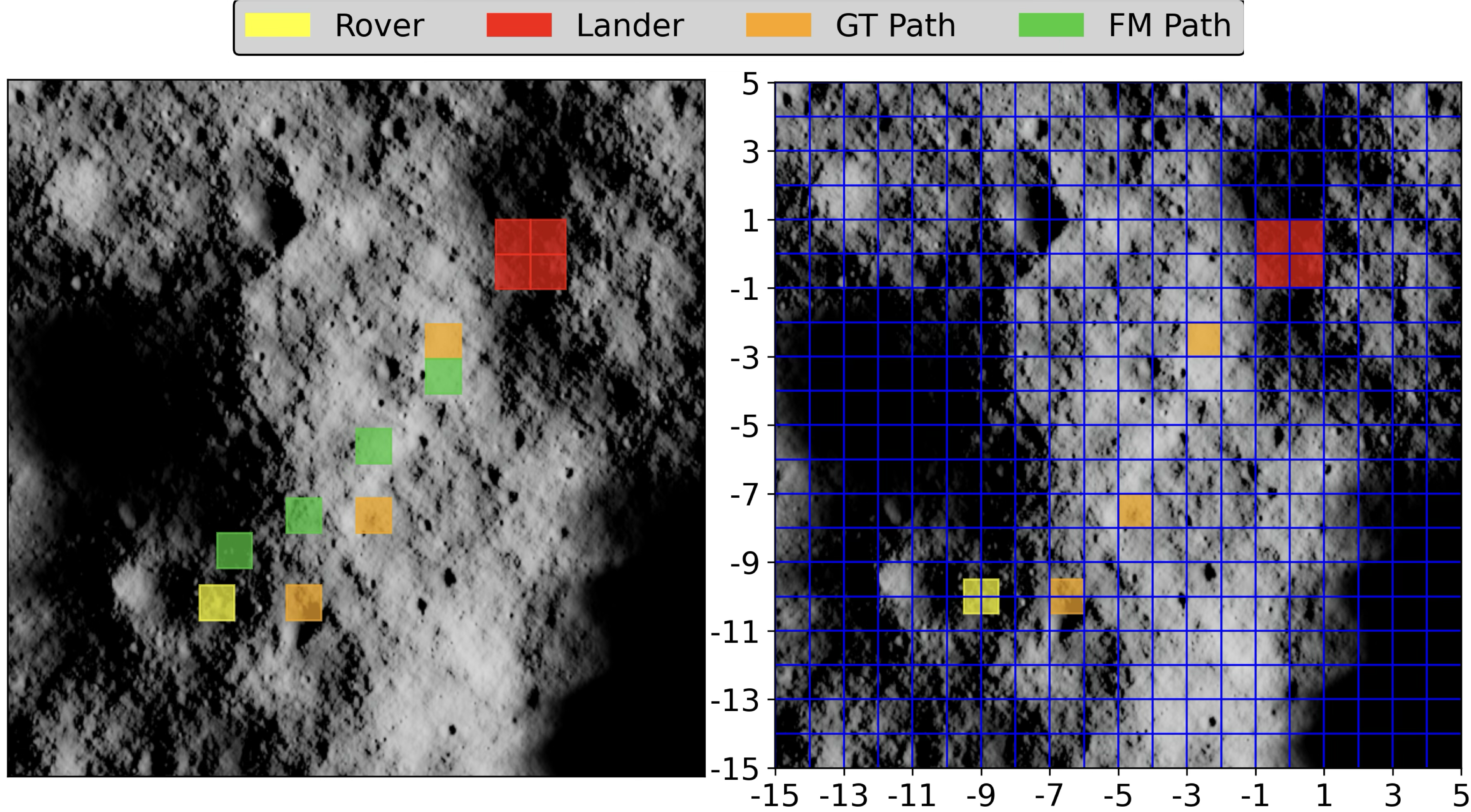}
        \caption{(Left) A top-down visualization showing the ground team’s (i.e., GT) proposed path (orange) alongside GPT-4o’s suggested alternative path (green). (Right) The top-down input image provided to GPT-4o, featuring a superimposed spatial grid to support the FM’s geometric reasoning.}
        \label{fig-lun-sim:top-down-generation-2}
    \end{subfigure}
    \caption{Large obstacles and structural threats scenario. (a) Front camera image. (b) Scenario visualizations.}
    \label{fig-lun-sim:scenario-2}
\end{figure}

\begin{tcolorbox}[colback=gray!40, colframe=gray!40, sharp corners]
{\fontfamily{cmtt}\selectfont 
FM Output
\newline
\newline
SCENE ANALYSIS:\newline
Using the three onboard camera images (Front, Front Left, and Front Right) alongside the top-down view, I have analyzed the observed terrain and conditions:\newline
\newline
**Front Camera**:\newline
- Terrain Features: A large rock is visible on the right half of the image. It denotes a significant obstacle.\newline
- Lighting Conditions: There are distinct shadows cast by the rock and the rover, which could obscure minor features or terrain details in its vicinity.\newline
- Navigation-Relevant Features: The presence of the rock on the right implies the rover should avoid moving significantly to the right to prevent collision.\newline
\newline
**Front Left Camera:**\newline
- Terrain Features: Similar to the front camera, the rock is clearly visible, occupying the center and right parts of the image.\newline
- Lighting Conditions: Shadows are apparent, especially around the rock, indicating potential challenges in judging distances.\newline
- Navigation-Relevant Feature: This observation corroborates the significant hazard that the rock presents. The terrain appears flatter to the left, presenting a clearer path.\newline
CONTINUED...
}
\end{tcolorbox}

\begin{tcolorbox}[colback=gray!40, colframe=gray!40, sharp corners]
{\fontfamily{cmtt}\selectfont 
...CONTINUED
\newline
**Front Right Camera:**\newline
- Terrain Features: The image shows the same large rock more prominently toward the left; additional smaller rocks are scattered throughout.\newline
- Lighting Conditions: Shadows again play a crucial role, suggesting that caution should be exercised in areas shadowed by the large rock.\newline
- Navigation-Relevant Features: Risks of minor obstacles due to smaller scattered rocks that may impede traversal.\newline
\newline
**Top-Down View:**\newline
- Ambient Details: The rover’s current position at (-9.5, -9.5) is accurately depicted alongside the lander at (0, 0).\newline
- Shadowed Areas: Shadowed regions are mainly towards the southwest quadrant of the top-down view, indicating limited visibility and terrain details in those areas.\newline
\newline
PLAN EVALUATION\newline

**Safety:**\newline
1. First Leg (Current location to -7, -9.5): Moving right from the current location could lead the rover into the large rock visible in the front and left camera views, creating a hazard.\newline
2. Second Leg (-7, -9.5 to -5, -7): This trajectory passes through relatively open areas.\newline
3. Third Leg (-5, -7 to -3, -2): No immediate hazards are visible along this path.\newline
4. Final Leg (-3, -2 to 0, 0): Appears clear without significant obstacles.\newline

**Feasibility:**\newline
- First Leg: Direct movement to the first waypoint is obstructed by a large rock; infeasible.\newline
- Remaining Legs: Feasible if clear paths are maintained with current imaging.\newline

**Efficiency:**\newline
- Current Path: Generally direct but necessitates unnecessary adjustments to the initial movement leftward.\newline
\newline
ALTERNATIVE SUGGESTIONS\newline

Due to the risk presented by the large rock, I propose an alternative path with clear and safe waypoints:\newline

Alternative Waypoints:\newline
1. (-9, -8) - Move left from the current position to bypass the large rock.\newline
2. (-7, -7) - Continue moving rightwards but upwards to match initial path direction.\newline
3. (-5, -5) - Adjust rightward to continue ahead cleanly.\newline
4. (-3, -3) - Single-step trajectory to provide smooth motion.\newline
5. (0, 0) - Direct final approach to the lander.\newline
\newline
Justification:\newline
This alternative path navigates around the major obstacle directly in front of the rover, utilizing clearer terrains visible from the onboard camera feeds. By maintaining a 2m distance from mapped hazards and avoiding shadowed areas, this plan ensures the rover travels safely and feasibly towards the lander while optimizing energy efficiency.
}
\end{tcolorbox}

\subsection{Low-visibility regions}

\textbf{Outputs and visualizations:}
\begin{figure}[ht!]
    \centering
    \begin{subfigure}{\linewidth}
        \centering
        \includegraphics[width=\linewidth]{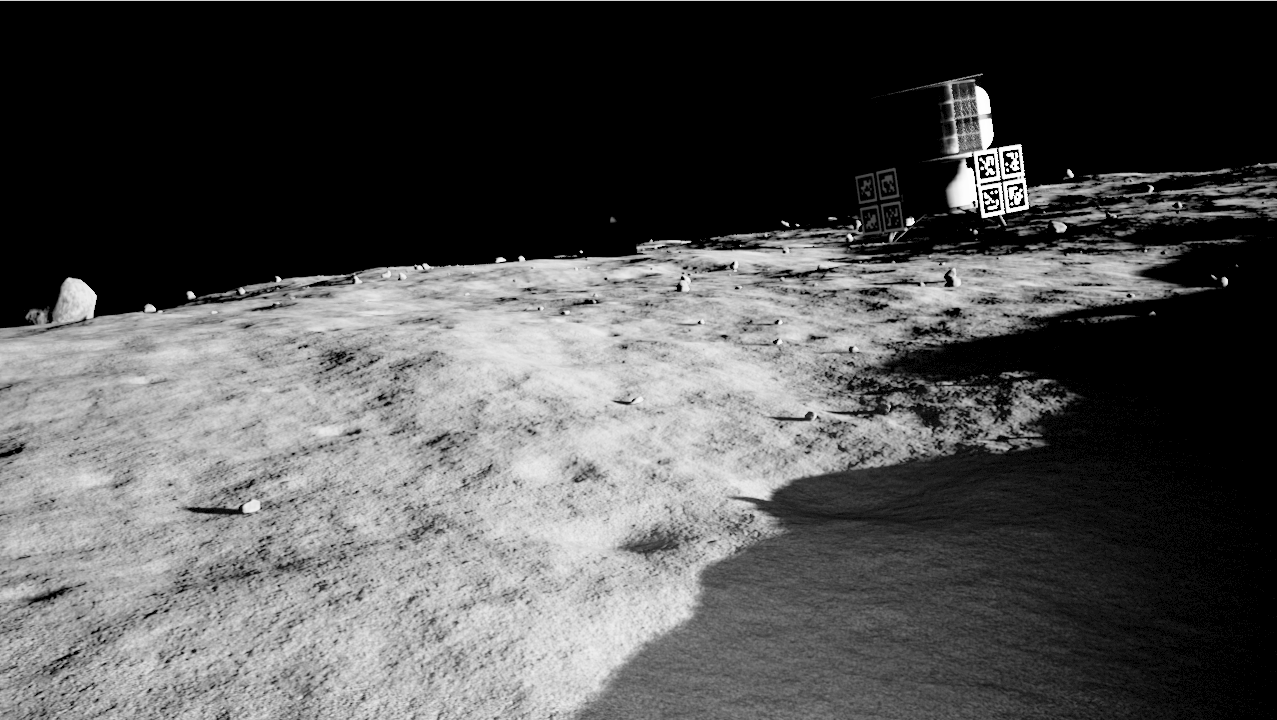}
        \caption{Onboard image provided to GPT-4.}
        \label{fig-lun-sim:front-scenario-3}
        \vspace{1.25em}
    \end{subfigure}
    \vspace{0.25em}
    \begin{subfigure}{\linewidth}
        \centering
        \includegraphics[width=\linewidth]{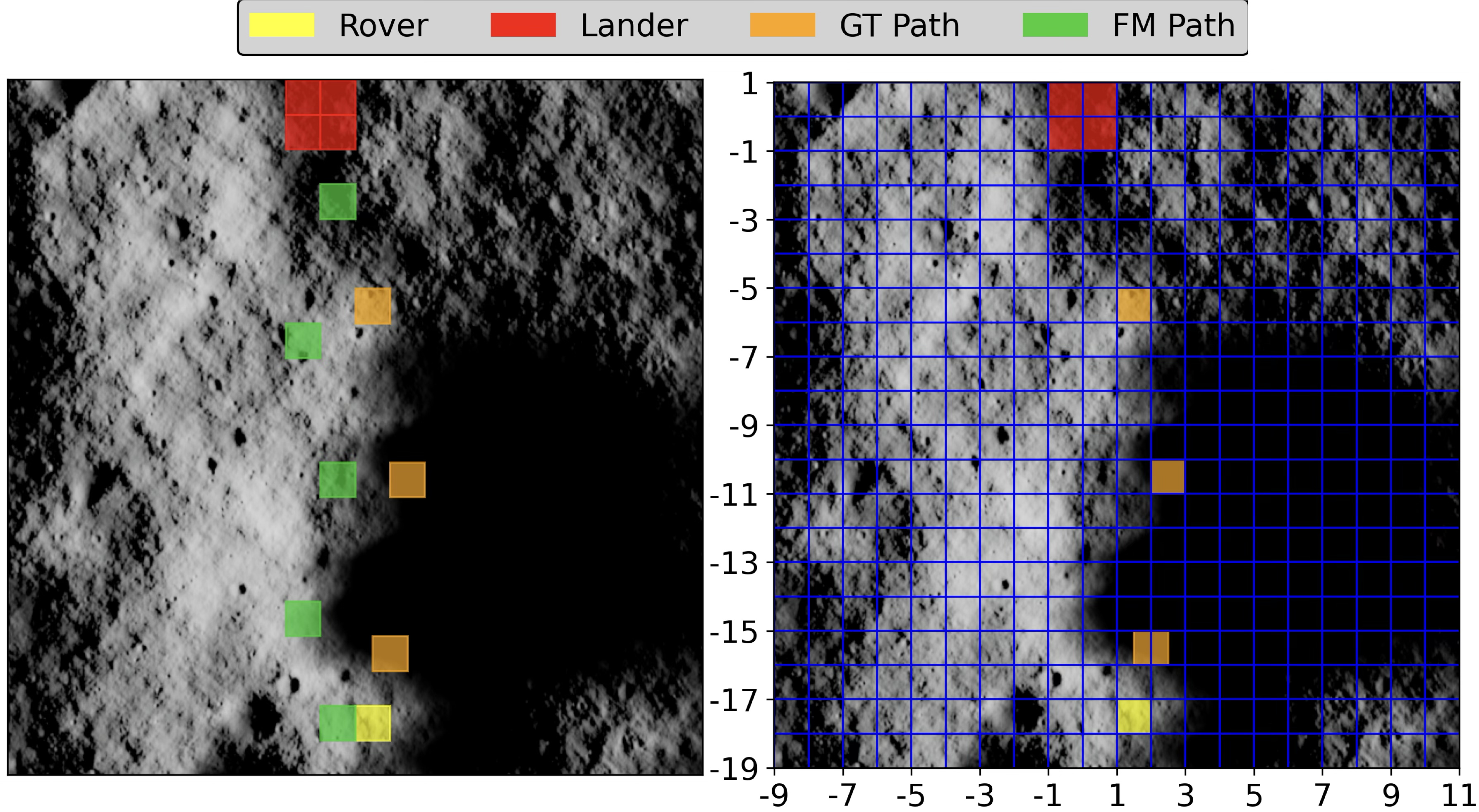}
        \caption{(Left) A top-down visualization showing the ground team’s (i.e., GT) proposed path (orange) alongside GPT-4o’s suggested alternative path (green). (Right) The top-down input image provided to GPT-4o, featuring a superimposed spatial grid to support the FM’s geometric reasoning.}
        \label{fig-lun-sim:top-down-generation-3}
    \end{subfigure}
    \caption{Low-visibility regions scenario. (a) Front camera image. (b) Scenario visualizations.}
    \label{fig-lun-sim:scenario-3}
\end{figure}

\begin{tcolorbox}[colback=gray!40, colframe=gray!40, sharp corners]
{\fontfamily{cmtt}\selectfont 
FM Output
\newline
\newline
SCENE ANALYSIS\newline

**Terrain Features:**\newline
- Front Camera: \newline
 - The terrain appears to be moderately flat with a slight upward slope. \newline
 - Several small to medium-sized rocks are scattered across the surface.\newline
 - A large lander is visible in the distance.\newline
- Front Left Camera: \newline
 - Similar terrain features as the front camera, with scattered rocks.\newline
 - The lander is visible to the right of the image.\newline
 - Shadows from terrain features or the lander itself are present towards the lower right.\newline
- Front Right Camera: \newline
 - Continuation of the terrain seen in the front left camera.\newline
 - The shadowed area in the foreground suggests a possible depression or slope.\newline

Lighting Conditions:\newline
- There are significant shadows from the lander and terrain features, primarily on the right side. \newline
- These shadows obscure parts of the terrain, which can mask potential hazards.\newline
- The areas to the left appear well-lit and more straightforward to navigate.\newline
CONTINUED...
}
\end{tcolorbox}

\begin{tcolorbox}[colback=gray!40, colframe=gray!40, sharp corners]
{\fontfamily{cmtt}\selectfont 
...CONTINUED\newline
Navigation-Relevant Features:\newline
- The immediate path forward seems clear with shallow slopes and scattered small rocks.\newline
- Shadowed areas present a challenge, particularly in identifying obstacles or slopes.\newline
\newline
PLAN EVALUATION\newline

**Safety:**\newline
- The path proposed by the ground team (heading to the right) starts at (1.5, -15) and passes through (2, -10), (1, -5), and finally (0, 0).
- Hazard Identification:\newline
- The path closely skirts the edge of significant shadowed areas which may obscure hazards.\newline
- The waypoint at (1.5, -15) and following segments through (2, -10) to (1, -5) likely encounter shadowed areas where hidden slopes or rocks are a risk.\newline

**Feasibility:**\newline
- Traversing through partially shadowed and rocky terrain may challenge the rover's mobility.\newline
- The rover's maneuverability near the lander without detailed assessment of the obstacles could lead to unexpected issues.\newline

**Efficiency:**\newline
- The proposed path follows a somewhat straight route but navigates through challenging terrain, potentially increasing energy consumption to avoid or climb obstacles.\newline
\newline
ALTERNATIVE SUGGESTIONS\newline
1. (0.5, -17)\newline
2. (0.5, -14)\newline
3. (0, -11)\newline
4. (0, 0)\newline

Path Description:\newline
1. Initial Movement:\newline
   - Move left (southwest) to (0.5, -17). This avoids the large shadowed area and stays mostly within the well-illuminated zone observed in the onboard camera images.\newline
   
2. Northwest Trajectory:\newline
   - Move northwest to (0.5, -14). Transition from well-lit area, ensuring to avoid major rocks or slope visible in the onboard cameras.\newline
   
3. Continue Northwest:\newline
   - Proceed to (0, -11). This waypoint stays within the clear zones identified with fewer obstructions, with minimal shadowed areas.\newline
   
4. Final Approach to Lander:\newline
   - Move towards the central position (0, 0), ensuring steady and clear path as initially visualized.\newline
\newline
Justification:\newline
- Safety:\newline
  - This path minimizes the number of shadowed areas the rover must navigate through, reducing the risk of encountering hidden obstacles.\newline
  - Maintains a safe distance from the edge of shadowed areas, allowing better visibility and safer traversal.\newline
- Feasibility:\newline
  - By avoiding large shadowed areas and opting for well-lit paths, the rover can better navigate through clear terrain without unnecessary complications.\newline
  - Additionally, this path allows safer conditions by avoiding rapid elevation changes or rocky outcrops directly in its initial straight path.\newline
- Efficiency:\newline
  - Avoiding rugged shadow-covered terrain ensures lower energy consumption by not forcing the rover to take detours around obstacles it might detect too late in darker sections.
}
\end{tcolorbox}







\end{document}